\documentclass{article} 
\usepackage{iclr2027_conference,times}

\usepackage{amsmath,amsfonts,bm}

\def\eqref#1{equation~\ref{#1}}

\def\1{\bm{1}}

\DeclareMathAlphabet{\mathsfit}{\encodingdefault}{\sfdefault}{m}{sl}
\SetMathAlphabet{\mathsfit}{bold}{\encodingdefault}{\sfdefault}{bx}{n}

\newcommand{\methodname}{\textsc{TabMF}}          

\newcommand{\dataset}{\mathcal{D}}
\newcommand{\encoder}{f_{\theta}}
\newcommand{\projector}{g_{\phi}}
\newcommand{\representation}{\mathbf{u}}
\newcommand{\embedding}{\mathbf{z}}

\usepackage{hyperref}
\usepackage{url}
\usepackage[capitalize,noabbrev]{cleveref}
\usepackage{graphicx}
\usepackage{booktabs}
\usepackage{multirow}
\usepackage{xcolor}
\usepackage{amsmath,amssymb}
\usepackage{subcaption}
\usepackage{tikz}                    
\usetikzlibrary{arrows.meta,calc}    
\usepackage{fontawesome5}            

\graphicspath{{figures/}}

\title{\methodname{}: Learning Dataset Representations with Tabular Foundation Models}

\author{
\makebox[\textwidth][c]{%
\begin{tabular}{c}
\textbf{Rafiq Kamel}\textsuperscript{1,}\thanks{Correspondence: \texttt{rafiq.kamel@utn.de}} \quad
\textbf{Nicolas Atienza}\textsuperscript{2} \quad
\textbf{Aziz Fouché}\textsuperscript{2} \\
\textbf{Nicola Cancedda}\textsuperscript{2} \quad
\textbf{Josif Grabocka}\textsuperscript{1} \\[0.5em]
\textsuperscript{1}University of Technology Nuremberg (UTN)
\qquad
\textsuperscript{2}Neuralk
\end{tabular}%
}
}

\iclrfinalcopy 

\begin{document}

\maketitle

\begin{abstract}
Representations of supervised datasets support higher-level machine-learning tasks such as dataset retrieval, and distribution-shift monitoring.
However, existing approaches typically rely either on hand-crafted meta-features or on dataset encoders trained specifically for representation learning.

We introduce \methodname{}, a framework that repurposes pretrained tabular foundation models (TFMs) as encoders of supervised datasets.
\methodname{} equips a predictive TFM with an architecture-aware dataset-level readout and adapts the model end-to-end using contrastive post-training on stochastic views of synthetic supervised datasets.


We instantiate \methodname{} on TabICLv2 and TabPFN-2.5 and evaluate the resulting representations across six tasks on unseen real datasets.
Simple frozen TFM readouts yield competitive representations, indicating that predictive pretraining already induces substantial dataset-level structure.
The two \methodname{} encoders obtain the strongest aggregate performance among the evaluated learned and classical dataset representations, with particularly strong results in dataset retrieval, compact benchmark construction, and distribution-shift severity assessment. 
\end{abstract}
\section{Introduction}
\label{sec:intro}

Many machine-learning problems require reasoning about an entire dataset rather than individual observations.
A useful representation of a supervised dataset can support dataset retrieval, benchmark construction, distribution-shift monitoring, transfer of model or hyperparameter knowledge, and identification of rare or atypical datasets.
Ideally, such a representation should map a previously unseen dataset to a fixed-dimensional vector whose geometry can be reused across different dataset-level tasks.

Existing approaches largely rely on hand-crafted meta-features or dedicated dataset encoders.
Meta-features summarize a dataset through a predefined collection of statistics~\citep{alcobaca2020pymfe} and underpin algorithm selection~\citep{mcelfresh2023tabzilla,zhao2021metaod} and zero-shot AutoML~\citep{ozturk2022zap,arango2024quicktune}, but their representation is fixed by design and cannot adapt to new notions of dataset similarity.
Learned alternatives such as Dataset2Vec~\citep{jomaa2021dataset2vec} and the
Neural Statistician~\citep{edwards2017neural} learn dedicated dataset encoders
from scratch on collections of datasets, without leveraging the representations
acquired by large-scale predictive pretraining.
This raises a more general question: can the representations learned through large-scale predictive pretraining be repurposed into reusable representations of entire supervised datasets?

We answer this question with \methodname{}, a foundation model for supervised dataset representation.
Our key idea is to repurpose pretrained tabular foundation models (TFMs) such as TabPFN~\citep{hollmann2025tabpfn,grinsztajn2025tabpfn} and TabICL~\citep{qu2025tabicl,qu2026tabiclv2betterfasterscalable}.
These models already process an entire labeled table jointly and learn rich internal representations through predictive pretraining, but are optimized for instance-level prediction rather than for representing the supervised dataset itself.
\methodname{} equips a pretrained TFM with an architecture-aware dataset-level readout and adapts the entire model end-to-end with a contrastive objective over stochastic views of synthetic supervised datasets.
The views vary sampled observations, observed variables, and label-preserving feature transformations, encouraging a representation that is stable across different realizations of the same supervised dataset.
We instantiate this general recipe on both TabICLv2 and TabPFN-2.5 using readouts tailored to their internal architectures (\cref{sec:method:encoders}).
At inference time, \methodname{} maps one supervised dataset to one fixed-dimensional vector in a single forward pass.

A surprising finding of our study is that predictive pretraining alone already induces useful dataset-level representations.
Simply mean-pooling internal states of frozen pretrained TFMs yields strong dataset embeddings without any dataset-level adaptation.
In particular, frozen TFM representations outperform dedicated learned dataset encoders on retrieval and remain competitive with learned encoders and classical meta-features on dataset-property prediction (\cref{sec:results:pretrained}).
This reveals an emergent dataset-level representation capability in predictive TFMs.

We evaluate frozen and adapted representations on real datasets unseen during training.
\methodname{} achieves near-perfect real-data view discrimination and substantially improves retrieval robustness under a broad suite of 19 dataset transformations, with particularly large gains under severe feature subsampling.
Beyond retrieval, the learned geometry supports higher-level applications including compact benchmark construction and distribution-shift severity assessment.
Across the six primary evaluation tasks, the two \methodname{} encoders obtain the two strongest aggregate ranks among the evaluated learned and classical dataset representations (\cref{sec:results}).

Our contributions are:
\begin{itemize}
    \item \textbf{\methodname{}.}
    We introduce an end-to-end framework that repurposes pretrained TFMs as zero-shot encoders of supervised datasets through architecture-aware readouts and contrastive post-training on synthetic datasets.
    At inference time, \methodname{} maps a previously unseen supervised dataset directly to a fixed-dimensional representation in a single forward pass (\cref{sec:method}).

    \item \textbf{Emergent dataset representations in predictive TFMs.}
    We show that predictive pretraining alone already induces strong dataset-level representations.
    Simple frozen readouts outperform dedicated learned dataset encoders on retrieval and remain competitive on dataset-property prediction (\cref{sec:results:pretrained}).

    \item \textbf{An extensive evaluation of dataset representations.}
    We compare frozen and adapted TFM representations with learned dataset encoders and classical meta-features across dataset-property prediction, robust retrieval, compact benchmark construction, and distribution-shift assessment.
    The two \methodname{} encoders achieve the strongest aggregate performance across the six-task primary suite (\cref{sec:results}).
\end{itemize}

\begin{figure}[t]
\begin{center}
\input{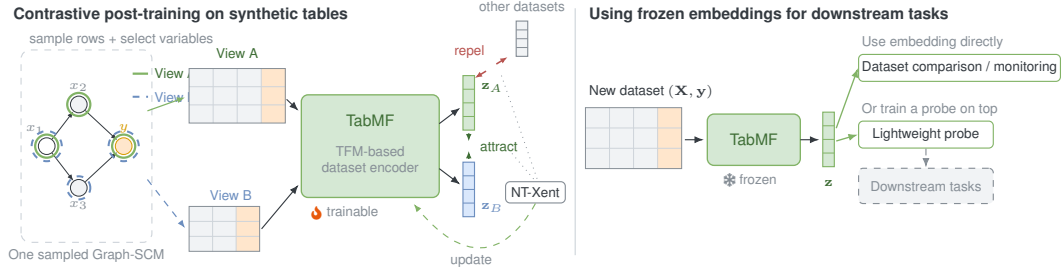}
\end{center}
\caption{
\textbf{\methodname{} overview.}
\textbf{Left:} A synthetic supervised dataset is sampled from one Graph-SCM, and two stochastic views vary the sampled observations, observed variables, and label-preserving feature transformations while sharing the same designated target.
\methodname{} encodes both views and is trained contrastively to bring their representations together while separating representations of other datasets.
\textbf{Right:} At inference time, a frozen \methodname{} encoder maps a new supervised dataset $(X,y)$ to a fixed-dimensional representation that can be used directly for dataset comparison and monitoring or as input to lightweight downstream probes.
}
\label{fig:overview}
\end{figure}
\section{Related Work}
\label{sec:related}

\textbf{Meta-features and learned dataset encoders.}
Hand-crafted meta-features summarize a dataset with fixed statistics~\citep{alcobaca2020pymfe} and underpin algorithm selection~\citep{mcelfresh2023tabzilla}, outlier-model selection~\citep{zhao2021metaod}, and zero-shot AutoML~\citep{ozturk2022zap,arango2024quicktune}.
As fixed descriptors, they cannot adapt their representation to the task consuming it, and \citet{herre2026metafeatures} report limited ability to explain performance differences between modern tabular foundation models.
Some meta-feature families also use explicit target statistics or fit predictive models to the dataset, which can overlap with downstream properties defined through predictive behavior; we control this explicitly for epiplexity prediction (\cref{app:results:pymfe}).
Learned alternatives train dataset encoders from scratch: Dataset2Vec~\citep{jomaa2021dataset2vec} distinguishes batches of the same versus different real datasets with a hierarchical set network, and the Neural Statistician~\citep{edwards2017neural} learns a variational summary of a set of exchangeable observations.
\methodname{} differs in starting from a pretrained TFM rather than a randomly
initialized set network, and in using synthetic supervised datasets for
dataset-level post-training; both baselines are included in our primary
comparison (\cref{sec:results}).
Task2Vec~\citep{achille2019task2vec} embeds tasks via Fisher information of a probe network, which requires fitting the probe on every new dataset; our encoders require only a forward pass.
\citet{kunz2026statistical} align numeric tables through their statistics, without learning and without supervised-task structure.

\textbf{Tabular foundation models and their representations.}
TabPFN~\citep{hollmann2025tabpfn,grinsztajn2025tabpfn} and TabICL~\citep{qu2025tabicl,qu2026tabiclv2betterfasterscalable} are Transformers pretrained on synthetic data for in-context label prediction; TabDPT~\citep{ma2025tabdpt} trains on real data, and CARTE~\citep{kim2024carte} incorporates column-name semantics.
Although these models are designed for prediction, their internal representations have also been reused: \citet{ye2025closerlook} study frozen TabPFN~v2 row-level features, while \citet{regenwetter2026engineering} pool TabPFN-2.5 activations into dataset descriptors for curating synthetic pretraining tasks.
We repurpose TabICLv2 and TabPFN-2.5 as encoders of supervised datasets using architecture-aware readouts and contrastive dataset-level post-training (\cref{sec:method:encoders}), and separately evaluate what their frozen readouts already capture (\cref{sec:results:pretrained}).

\textbf{Contrastive learning and table embeddings.}
Contrastive methods on tabular data corrupt features~\citep{bahri2022scarf} or subset them~\citep{ucar2021subtab} to construct positive views of individual rows, following the instance-level augmentation paradigm of SimCLR~\citep{chen2020simpleframeworkcontrastivelearning}.
\methodname{} instead applies contrastive learning at the dataset level: positive pairs are stochastic views of the same supervised dataset that can differ in their sampled observations, observed variables, and label-preserving feature transformations (\cref{sec:method:contrastive}).

A separate line of work learns table representations for content-based retrieval and data discovery, including HyTrel~\citep{chen2023hytrel}, the TEmBed benchmark~\citep{vogel2026tembed,poostforoushan2026tembedt}, and generalist tabular embeddings~\citep{qiang2026tabembed}.
This setting is distinct from ours: \methodname{} represents a supervised dataset $(X,y)$, with the designated target forming part of the represented object, rather than a target-agnostic table.
We nevertheless compare against HyTrel and report TEmBed results separately in Appendix~\ref{app:results:transfer} to examine how supervised dataset representations transfer to content-oriented table tasks.
\section{\methodname{}}
\label{sec:method}

Let $\dataset=(\mathbf{X},\mathbf{y})$ denote a supervised classification
dataset with $T$ observations, $H$ features, and $C$ classes, where
$\mathbf{X}\in\mathbb{R}^{T\times H}$ and
$\mathbf{y}\in\{0,\ldots,C-1\}^{T}$.
We seek an encoder $\encoder$ that maps a previously unseen supervised dataset
to a fixed-dimensional representation
$\embedding=\encoder(\dataset)\in\mathbb{R}^{d_z}$.
The designated target $\mathbf{y}$ is part of the represented object, so the
embedding is intended to capture predictive structure between features and
target rather than target-agnostic table content.

\methodname{} repurposes a pretrained predictive tabular foundation model
(TFM) as such a dataset encoder.
The general recipe combines three components:
(i) a pretrained TFM backbone that contextualizes an entire labeled dataset,
(ii) an architecture-aware readout that reduces its internal states to a
single dataset representation, and
(iii) contrastive post-training over stochastic views of synthetic supervised
datasets.
The complete encoder is fully differentiable, allowing the backbone,
readout, and projection network to be optimized jointly.
At inference time, a dataset is mapped directly to a fixed-dimensional
embedding in a single forward pass.

We instantiate this recipe on TabICLv2
\citep{qu2026tabiclv2betterfasterscalable} and TabPFN-2.5
\citep{grinsztajn2025tabpfn}.
We refer to the resulting models as \methodname{}-ICL and
\methodname{}-PFN, respectively, for brevity.
The two models organize their internal representations differently:
TabICLv2 reduces observations to fixed-dimensional row states before applying
cross-row in-context attention, whereas TabPFN-2.5 maintains contextual
representations jointly over observations and feature groups.
We therefore use a different dataset-level readout for each backbone
(\cref{sec:method:encoders}).

We also evaluate dataset-level readouts of the frozen backbones to measure
what predictive pretraining provides before contrastive post-training; implementation details are given in
Appendix~\ref{app:raw-tfm-representations}.

\subsection{Contrastive Dataset-Level Adaptation}
\label{sec:method:contrastive}

\methodname{} shapes dataset representations by encouraging consistency across
different views of the same supervised dataset.
We draw $B$ synthetic supervised datasets
$\dataset_i=(\mathbf{X}_i,\mathbf{y}_i)\sim p_{\mathrm{syn}}$,
$i=1,\ldots,B$, from the graph-structured prior of TabICLv2
\citep{qu2026tabiclv2betterfasterscalable}.

For each dataset, we independently construct two variable-sized views.
Rows are resampled with replacement, and features are subsampled without
replacement.
Subsets of the selected features may also undergo noise or Fourier
transformations, while their labels remain unchanged.
Let $\mathcal{I}_i^v$ and $\mathcal{J}_i^v$ denote the sampled row and feature
indices, and let $a_i^v$ denote the feature transformation.
For $v\in\{A,B\}$, the resulting dataset view is
\begin{equation}
    \dataset_i^v
    =
    \left(
        a_i^v\!\left(
            \mathbf{X}_i[
                \mathcal{I}_i^v,
                \mathcal{J}_i^v
            ]
        \right),
        \mathbf{y}_i[\mathcal{I}_i^v]
    \right).
    \label{eq:dataset-view}
\end{equation}
The views can differ in their observations, observed variables, feature
values, and sizes while retaining the same designated prediction target.
Sampling ranges and transformation parameters are specified in \cref{sec:exp}.

The shared encoder maps the views to
$\embedding_i^A=\encoder(\dataset_i^A)$ and
$\embedding_i^B=\encoder(\dataset_i^B)$.
The two views of a source dataset form a positive pair; views of different
source datasets in the batch form negatives.
Positive pairs are defined by their shared source dataset, rather than by
matching parameters of the synthetic generator, which are not provided to
the encoder.

The architecture-aware readout produces a dataset representation
$\representation$.
A projection network consisting of LayerNorm followed by a two-layer GELU
MLP maps it to the normalized embedding
$\embedding
=\projector(\representation)/
\lVert\projector(\representation)\rVert_2$.

We train \methodname{} with the normalized temperature-scaled cross-entropy
objective (NT-Xent)
\citep{chen2020simpleframeworkcontrastivelearning}.
For an anchor $\embedding_i^v$, let $\embedding_i^{\bar v}$ denote the
embedding of its other view.
The loss is
\begin{equation}
    \ell_i^v
    =
    -\log
    \frac{
        \exp\!\left(
            \operatorname{sim}
            (\embedding_i^v,\embedding_i^{\bar v})/\tau
        \right)
    }{
        \displaystyle
        \sum_{(j,w)\neq(i,v)}
        \exp\!\left(
            \operatorname{sim}
            (\embedding_i^v,\embedding_j^w)/\tau
        \right)
    },
    \label{eq:ntxent}
\end{equation}
where $j\in\{1,\ldots,B\}$ indexes source datasets and
$w\in\{A,B\}$ indexes views.
The denominator includes all $2B$ views except the anchor;
$\operatorname{sim}(\mathbf{a},\mathbf{b})=\mathbf{a}^{\top}\mathbf{b}$
for normalized embeddings, and $\tau$ is the temperature.
We average the loss over both views of all $B$ datasets.

\subsection{Architecture-Aware \methodname{} Encoders}
\label{sec:method:encoders}

Both encoders use the contrastive objective described above, but their dataset-level readouts follow the structure of their respective backbones. \Cref{fig:encoders} in Appendix~\ref{sec:illustrationaggregation} illustrates the two designs.

For the \methodname{} variant with the TabICLv2 backbone, we combine each
row state with its target embedding,
$\widetilde{\mathbf{r}}_t
=\mathbf{r}_t+\operatorname{YEnc}(y_t)$, and append a learned summary token with no associated input observation or target label $\mathbf{R}=
[\widetilde{\mathbf{r}}_1,\ldots,\widetilde{\mathbf{r}}_T,
\mathbf{s}_{\mathrm{sum}}]$. We take its final state as the dataset representation:
\begin{equation}
    \representation_{\mathrm{TabICL}}
    =
    \operatorname{LN}_{\mathrm{ICL}}
    \left(
        \operatorname{TF}_{\mathrm{ICL}}^{(T)}
        \left(\mathbf{R}\right)
    \right)_{T+1}.
    \label{eq:tabicl-summary}
\end{equation}

For the variant with the TabPFN-2.5 backbone, we denote the feature-group states as $\mathbf{F}\in\mathbb{R}^{T\times M\times d_h}$,
where $M$ is the number of feature-group positions and $d_h$ is their
hidden dimension. We average each group over observations,
$\bar{\mathbf{f}}_m
=T^{-1}\sum_{t=1}^{T}\mathbf{F}_{t,m}$,
then apply a learned Transformer across the resulting group states,
$\widetilde{\mathbf{F}}
=\operatorname{TF}_{\mathrm{col}}(\bar{\mathbf{F}})$,
where $\bar{\mathbf{F}}$ collects
$\bar{\mathbf{f}}_1,\ldots,\bar{\mathbf{f}}_M$. Writing $\mathcal{M}_{\mathrm{active}}$ for the set of non-padded groups,
we obtain the dataset representation as
\begin{equation}
    \representation_{\mathrm{TabPFN}}
    =
    \frac{1}{|\mathcal{M}_{\mathrm{active}}|}
    \sum_{m\in\mathcal{M}_{\mathrm{active}}}
    \widetilde{\mathbf{f}}_m.
    \label{eq:tabpfn-column-readout}
\end{equation}

\section{Experimental Setup}
\label{sec:exp}

\subsection{Training and Models}

\noindent\textbf{Training.}
We generate synthetic datasets on the fly using the TabICLv2 Graph-SCM prior
\citep{qu2026tabiclv2betterfasterscalable}, with 5,000--50,000 rows,
5--100 features, and at most ten classes, retaining the remaining prior
settings.
Each dataset yields two stochastic views as described in
\cref{sec:method:contrastive}, containing 500--5,000 rows and 5--50 features
for \methodname{}-ICL, or 200--1,500 rows and 5--30 features for
\methodname{}-PFN, capped at the source dimensions.
The smaller \methodname{}-PFN views reflect its higher memory requirements.
Training views additionally apply noise and Fourier transformations to 30\%
of the selected columns, with noise standard deviation $0.1$ and 16 Fourier
bins.

We jointly optimize the backbone, architecture-aware readout, and projection
network using AdamW with a global batch size of 64 datasets, a learning rate
of $10^{-5}$, zero weight decay, and a cosine learning-rate schedule with
5\% warmup.
The evaluated \methodname{}-ICL and \methodname{}-PFN checkpoints use 45,000
optimization steps.
Training uses two NVIDIA H100 GPUs per encoder and requires approximately
315 GPU-hours for \methodname{}-ICL and 273 GPU-hours for
\methodname{}-PFN.

\noindent\textbf{Encoders and baselines.}
We evaluate the summary-token and column-aware encoders defined in
\cref{sec:method:encoders}, instantiated with TabICLv2
\citep{qu2026tabiclv2betterfasterscalable} and TabPFN-2.5
\citep{grinsztajn2025tabpfn}.
TabICL ICL-mean and TabPFN mean-pooling provide frozen-backbone baselines;
comparisons with the adapted encoders therefore include changes in readout.
We also evaluate alternative TabICL readouts and random initialization,
alongside PyMFE~\citep{alcobaca2020pymfe},
Dataset2Vec~\citep{jomaa2021dataset2vec},
the Neural Statistician~\citep{edwards2017neural}, and
HyTrel~\citep{chen2023hytrel}.

\subsection{Evaluation Tasks}

We evaluate transfer from synthetic post-training datasets to previously
unseen real datasets across the following dataset-level tasks.

\noindent\textbf{Bootstrap view discrimination.}
On 191 TabBench datasets~\citep{neuralk2025tabench}, we test whether views
of the same dataset have higher cosine similarity than views of different
datasets.
Four views are generated per dataset by independently resampling rows and
subsampling features, with 500--5,000 rows and 5--50 features, capped at the
source dimensions.
ROC-AUC measures discrimination between same-dataset and cross-dataset view
pairs.
Per-dataset hard margins additionally measure separation from the closest
competing views.

\noindent\textbf{Dataset-property prediction.}
We probe embeddings for two properties on TabBench.
Domain discrimination predicts eight manually assigned domains with a
multiclass logistic-regression probe and reports macro one-vs-rest ROC-AUC.
Epiplexity measures the structural information that can be extracted from a
dataset by a learner of fixed computational capacity
\citep{finzi2026entropy}.
For epiplexity prediction, we use prequential estimates obtained with an
online logistic classifier as reference values.
We train a ridge regressor to predict these estimates from dataset embeddings
and evaluate Spearman correlation on held-out datasets. 
PyMFE uses a leakage-controlled feature set for epiplexity (Appendix~\ref{app:results:pymfe}).


Appendix~\ref{app:results:properties} specifies preprocessing, coverage, and
paired analyses.
The epiplexity estimator and additional probes are described in
Appendix~\ref{app:results:epiplexity}.

\noindent\textbf{Retrieval under transformations.}
A transformed dataset retrieves its source from a gallery of original
datasets by cosine similarity; encoder-specific coverage is detailed
in Appendix~\ref{app:results:transformations}.
The 19 transformations cover row and feature subsampling, recoding,
rescaling, missingness, noise, and quantization.
Mean recall@5 across transformations measures preservation of dataset
identity as the observed table changes.

\noindent\textbf{Compact benchmark selection.}
Following \citet{gusev2026benchmarking}, we select three datasets and measure
how closely their classifier ranking matches that of the full benchmark.
We use 144 classification datasets and ten default classifiers
from TabRepo \citep{salinas2024tabrepolargescalerepository}.
Each of 200 shared trials draws a candidate pool containing 80\% of the
population.
K-means selects a dataset representative from each cluster; cosine
farthest-first provides a second selector.
The metric is mean Spearman correlation with the full-population classifier
ranking (Appendix~\ref{app:results:benchmark-selection}).

\noindent\textbf{Distribution-shift severity.}
On nine TableShift tasks~\citep{gardner2023tableshift}, we test whether
ID--OOD embedding distances order shifts consistently with their published
mean accuracy drops across estimators.
Encoders receive the same label-stratified samples of at most 8,192 rows
from each official test split, and preprocessing is fitted on the training split.
We compute Spearman correlation between cosine distance and the mean
ID-minus-OOD accuracy gap, with one observation per task; the nine tasks,
severity targets, and sampled population sizes are reported in
Table~\ref{tab:tableshift-inputs}.
\section{Results}
\label{sec:results}

We evaluate whether \methodname{} provides a reusable representation of
previously unseen supervised datasets across several forms of dataset-level
reasoning.
We first summarize performance across the six primary evaluation tasks, then
examine robustness to alternative views of the same dataset and applications
of the learned geometry to compact benchmark construction and
distribution-shift assessment.
Finally, we analyze how much dataset-level structure is already present in
frozen predictive TFMs and examine the effect of encoder design choices.

We select one variant per backbone using bootstrap-view discrimination on 50
synthetic datasets generated with the TabICLv2 Graph-SCM prior
\citep{qu2026tabiclv2betterfasterscalable}.
The selected summary-token and column-aware encoders are used throughout;
Appendix~\ref{app:results:selection} gives the candidate comparison.
This synthetic benchmark is used only for variant selection and does not enter
the primary evaluation or aggregate ranking.
Table~\ref{tab:results-primary} summarizes the six primary evaluations on
real datasets.

\begin{table}[t]
\centering
\small
\setlength{\tabcolsep}{3pt}
\caption{Primary results across six evaluation tasks. Scores are real-data bootstrap AUC, domain macro-OVR AUC, epiplexity regression Spearman, mean retrieval recall@5, TabRepo three-dataset K-means ranking Spearman, and TableShift shift-severity Spearman. Encoders are ordered by their median rank across the six tasks. Higher scores and lower ranks are better; bold marks numerical leaders. Epiplexity uses available-case scores and leakage-controlled PyMFE features; paired common-population results are in Appendix~\ref{app:results:epiplexity}.}
\label{tab:results-primary}
\begin{tabular}{@{}lrrrrrrr@{}}
\toprule
& \multicolumn{4}{c}{TabBench} & TabRepo & TableShift & \\
\cmidrule(lr){2-5}\cmidrule(lr){6-6}\cmidrule(lr){7-7}
Encoder & Views & Domain & Epiplexity & Recall@5 & Selection & Severity & Rank $\downarrow$ \\
\midrule
\methodname{}-PFN & \textbf{0.9985} & 0.683 & 0.597 & \textbf{0.965} & \textbf{0.929} & 0.917 & \textbf{1.5} \\
\methodname{}-ICL & 0.9981 & \textbf{0.702} & 0.639 & 0.938 & 0.854 & \textbf{0.933} & 2.0 \\
TabICL ICL-mean & 0.9775 & 0.670 & \textbf{0.742} & 0.889 & 0.575 & 0.850 & 3.5 \\
PyMFE & 0.9812 & 0.661 & 0.106 & 0.931 & 0.709 & 0.733 & 4.0 \\
TabPFN mean-pooling & 0.9414 & 0.650 & 0.741 & 0.821 & 0.850 & 0.583 & 5.0 \\
Dataset2Vec & 0.9790 & 0.643 & 0.536 & 0.750 & 0.652 & 0.617 & 5.5 \\
Neural Statistician & 0.7917 & 0.554 & 0.239 & 0.677 & 0.738 & $-0.150$ & 7.0 \\
\bottomrule
\end{tabular}
\end{table}

\subsection{\methodname{} Provides Strong Dataset Representations}
\label{sec:results:overall}

Across the six primary evaluation tasks, the two \methodname{} variants obtain
the two strongest aggregate ranks, with median ranks of $1.5$ for
\methodname{}-PFN and $2.0$ for \methodname{}-ICL
(Table~\ref{tab:results-primary}).
A \methodname{} encoder is the numerical leader on five of the six individual
tasks, spanning real-data view discrimination, domain prediction, robust
retrieval, compact benchmark selection, and distribution-shift severity.
The exception is epiplexity prediction, where frozen TFM representations are
stronger under the primary linear probe; we return to this result in
\cref{sec:results:pretrained}.

These results indicate that the same learned representation can support
multiple dataset-level objectives rather than being useful only for the
contrastive task used during training. We additionally examine computational
scaling in Appendix~\ref{app:results:efficiency}, where \methodname{} produces
dataset representations substantially faster than classical meta-feature
extraction on large tables while adding little inference overhead relative
to the corresponding frozen TFM representations. We next examine the
strongest effects in more detail.

\subsection{\methodname{} Learns a Robust Geometry for Dataset Comparison}
\label{sec:results:robustness}

Both \methodname{} encoders transfer the separation learned from synthetic
training views to unseen real datasets.
On real-data bootstrap-view discrimination, \methodname{}-ICL and
\methodname{}-PFN achieve ROC-AUCs of $0.9981$ and $0.9985$, respectively
(Table~\ref{tab:results-primary}).
Because several baselines also obtain high AUC, we additionally evaluate the
harder per-dataset margin criterion.
Both \methodname{} variants obtain significantly larger hard margins
than all nine baselines in this comparison
(Table~\ref{tab:results-bootstrap-significance});
the margin construction and paired tests are detailed in
Appendix~\ref{app:results:selection}.

The advantage persists under substantially broader transformations.
When recall@5 is averaged across applicable transformations within each
dataset, both \methodname{} variants significantly outperform TabICL
ICL-mean, while \methodname{}-PFN additionally improves significantly over PyMFE (Appendix~\ref{app:results:transformations}). The clearest shared advantage occurs under severe feature removal: when only 25\% of features remain, recall@5 is $0.884$ and $0.937$ for \methodname{}-ICL and \methodname{}-PFN, compared with $0.653$ for TabICL ICL-mean and $0.728$ for PyMFE.


\begin{figure}[t]
  \centering
  \begin{subfigure}[b]{0.44\textwidth}
    \centering
    \includegraphics[width=\linewidth]{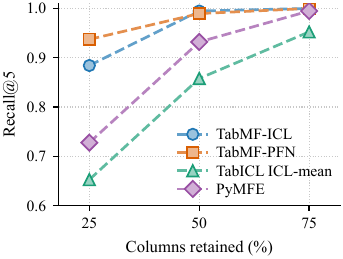}
    \caption{Recall@5 vs.\ fraction of columns retained.}
    \label{fig:robustness-cols}
  \end{subfigure}\hfill
  \begin{subfigure}[b]{0.53\textwidth}
    \centering
    \includegraphics[width=\linewidth]{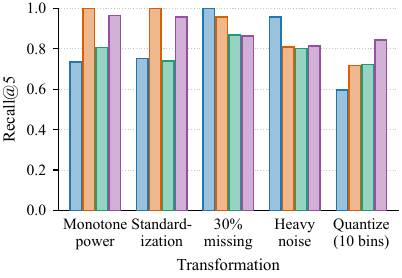}
    \caption{Recall@5 under value-level transformations.}
    \label{fig:robustness-transforms}
  \end{subfigure}
  \caption{Dataset identity remains recoverable with incomplete features, but robustness
    depends on the encoder and transformation.
    \textbf{(\subref{fig:robustness-cols})}~Both \textsc{TabMF} variants outperform ICL-mean and PyMFE
    when only 25\% of columns remain.
    \textbf{(\subref{fig:robustness-transforms})}~\textsc{TabMF}-PFN better tolerates monotone
    transformations and standardization, while \textsc{TabMF}-ICL better tolerates heavy
    noise and missingness; quantization limits both variants.
    Values are retrieval recall@5; the complete transformation results appear in
    Figure~\ref{fig:results-transforms-full}.}
  \label{fig:robustness}
\end{figure}

The two variants show different robustness profiles
(Figure~\ref{fig:robustness}).
\methodname{}-PFN better preserves dataset identity under monotone
transformations and standardization, whereas \methodname{}-ICL is more robust
to heavy noise and missingness.
Quantization remains a shared limitation, with both variants falling below
PyMFE.
Overall, \methodname{} substantially improves robustness to alternative views
of the same dataset, particularly when much of the original feature set is
unavailable.

\subsection{Compact Benchmark Selection Preserves Classifier Rankings}
\label{sec:results:benchmark-selection}

We next test whether the learned geometry can support a dataset-level decision
task that is not used during training.
At a three-dataset budget, \methodname{}-PFN achieves the strongest
classifier-ranking correlation on TabRepo, reaching $0.929$ with K-means
selection compared with $0.850$ for frozen TabPFN mean-pooling and $0.777$
for random selection (Table~\ref{tab:results-primary}).

The result is also consistent across the two evaluated selection rules at the
three-dataset budget. Under cosine farthest-first selection, \methodname{}-PFN
reaches 0.823, compared with 0.673 for frozen TabPFN mean-pooling and 0.777
for random selection. \methodname{}-ICL also exceeds random selection under
both selectors, reaching 0.854 with K-means and 0.815 with farthest-first.
The result is robust across nearby compact selection budgets:
for each $k\in\{2,3,4\}$, a \methodname{} variant obtains the highest
numerical score under both K-means and farthest-first selection.
The relative ordering of \methodname{}-ICL and \methodname{}-PFN varies
with the selection budget, and Appendix~\ref{app:results:benchmark-selection}
reports the full sensitivity analysis.

\subsection{Embedding Distances Track Distribution-Shift Severity}
\label{sec:results:tableshift}

The learned representation also provides a meaningful notion of distance
between related dataset distributions.
Across the nine TableShift tasks, the Spearman correlation between ID--OOD
embedding distance and the published mean accuracy drop is $0.933$ for
\methodname{}-ICL and $0.917$ for \methodname{}-PFN
(Table~\ref{tab:results-primary}).
The strongest baseline, TabICL ICL-mean, obtains $0.850$.

This evaluation complements transformation retrieval.
Retrieval rewards invariance to changes that should preserve dataset identity,
whereas TableShift rewards sensitivity to population changes associated with
deteriorating predictive performance.
The high observed correlations therefore indicate that the learned geometry
can be invariant to nuisance transformations while remaining sensitive to
consequential distribution shifts.
Because the evaluation contains only nine task-level observations, we do not
interpret the numerical ordering as evidence of pairwise superiority between
encoders; Appendix~\ref{app:results:statistics} summarizes the statistical
scope of the aggregate evaluation.

\subsection{Predictive TFMs Already Encode Dataset-Level Information}
\label{sec:results:pretrained}

Having established the performance of \methodname{}, we ask how much of this
dataset-level capability is already induced by predictive pretraining.
The answer is: a substantial part.
Without any dataset-level optimization, TabICL ICL-mean achieves a mean
retrieval recall@5 of $0.889$, outperforming Dataset2Vec ($0.750$) and the
Neural Statistician ($0.677$), while remaining close to PyMFE ($0.931$).
Frozen TFM representations also encode substantial dataset-property
information: TabICL ICL-mean and TabPFN mean-pooling obtain domain AUCs of
$0.670$ and $0.650$ and epiplexity correlations of $0.742$ and $0.741$,
respectively.

We use tuned linear probes for the primary property evaluations because they
provide a simple, controlled measure of how readily each property is accessible
from a representation.
Under these probes, dataset-level post-training does not uniformly improve
property recoverability.
Although \methodname{}-ICL obtains the highest observed domain AUC ($0.702$),
the paired analysis does not establish an advantage over the corresponding
frozen readouts (Appendix~\ref{app:results:tabbench}).
For continuous epiplexity prediction, both frozen encoders obtain higher
correlations than their \methodname{} counterparts, with three of the four
cross-comparisons significant after correction
(Appendix~\ref{app:results:epiplexity}).

These primary comparisons characterize linear accessibility rather than total
information content.
We therefore repeat both evaluations with substantially more expressive
nonlinear probes as a capacity check.
With a tuned RBF-and-polynomial SVM, \methodname{}-ICL obtains a higher
domain AUC than frozen TabICL ($0.700$ versus $0.644$), and the corresponding
paired per-dataset log-loss comparison is significant after Holm correction
($p=0.0278$ over the full 11-encoder comparison pool). The corresponding PFN
comparison remains unresolved (Appendix~\ref{app:results:tabbench}).
With a strengthened multi-kernel ridge probe for epiplexity, neither adapted
encoder is significantly separated from its corresponding frozen
representation under the same full-pool correction scope.
The complete epiplexity results are reported in
Appendix~\ref{app:results:epiplexity}.

Together, these results show that predictive TFM pretraining itself induces
useful representations of entire supervised datasets.
Dataset-level post-training strongly improves the geometry for robust dataset
comparison, but can also change how individual properties are organized in the
representation space.
The linear results should therefore be read as differences in readily
accessible structure; the nonlinear checks show that the apparent ties and
losses do not, by themselves, establish that the corresponding information is
absent from the adapted representations.

\subsection{Ablations}
\label{sec:results:ablations}

Table~\ref{tab:results-ablations} compares the evaluated initialization,
objective, and readout variants using mean retrieval recall@5.
Pretrained initialization substantially outperforms scratch initialization
for both backbones under matched architecture, readout, and NT-Xent training:
recall increases from $0.797$ to $0.938$ for the TabICL summary-token encoder
and from $0.587$ to $0.965$ for the TabPFN column-aware encoder.

Among the evaluated objectives, NT-Xent obtains the highest recall for both
selected architectures: $0.938$ versus $0.924$ for triplet and $0.883$ for
exponential distance with \methodname{}-ICL, and $0.965$ versus $0.962$ and
$0.920$, respectively, with \methodname{}-PFN.

For TabPFN-2.5, the column-aware readout with NT-Xent obtains $0.965$ recall,
compared with $0.935$ for the summary-token readout and $0.957$ for the pseudo-test variant.

\begin{table}[t]
\centering
\small
\setlength{\tabcolsep}{4pt}
\caption{Encoder design comparisons using mean retrieval recall@5.
Each block compares configurations within one backbone across
initialization, readout, and objective choices.}
\label{tab:results-ablations}
\begin{tabular}{@{}llllr@{}}
\toprule
Backbone & Readout & Initialization & Objective & Recall@5 \\
\midrule
TabICL & Summary-token & Pretrained & NT-Xent & \textbf{0.938} \\
TabICL & Summary-token & Scratch & NT-Xent & 0.797 \\
TabICL & Summary-token & Pretrained & Triplet & 0.924 \\
TabICL & Summary-token & Pretrained & Exp.\ distance & 0.883 \\
\midrule
TabPFN & Column-attention & Pretrained & NT-Xent & \textbf{0.965} \\
TabPFN & Column-attention & Scratch & NT-Xent & 0.587 \\
TabPFN & Summary-token & Pretrained & NT-Xent & 0.935 \\
TabPFN & Pseudo-test & Pretrained & NT-Xent & 0.957 \\
TabPFN & Column-attention & Pretrained & Triplet & 0.962 \\
TabPFN & Column-attention & Pretrained & Exp.\ distance & 0.920 \\
\bottomrule
\end{tabular}
\end{table}

\section{Discussion and Limitations}
\label{sec:discussion}

Our findings show that predictive TFMs can support reusable dataset
representations across several forms of dataset-level reasoning.
We deliberately focus on adapting existing pretrained TFMs with a limited
additional training budget.
Pretrained initialization outperforms training from scratch for both
backbones, supporting this practical reuse strategy.
The representations are explicitly supervised: changing the designated
target can change the representation even when the feature table is fixed.
Our evaluation covers classification datasets and two pretrained TFM
backbones; extending the framework to regression and other architectures
would further establish its generality.
Post-training uses the TabICLv2 Graph-SCM prior, and the resulting encoders
transfer to unseen real datasets.
Broader synthetic priors or mixtures of synthetic and real datasets offer
a direction for extending this coverage.
Different applications may also require different notions of dataset
similarity, motivating representations whose geometry can be conditioned
on the downstream objective.

\section{Conclusion}
\label{sec:conclusion}

We introduced \methodname{}, a framework for repurposing pretrained tabular
foundation models as encoders of entire supervised datasets.
Through architecture-aware dataset readouts and contrastive post-training on
synthetic dataset views, \methodname{} maps an unseen supervised dataset to a
fixed-dimensional representation in a single forward pass.
Across six primary evaluations on real datasets, the two \methodname{}
instantiations obtain the strongest aggregate performance among the evaluated
learned and classical dataset representations, with particularly strong
results in robust dataset comparison, compact benchmark construction, and
distribution-shift assessment.

We additionally find that predictive TFM pretraining alone already induces
useful dataset-level representations, even without explicit dataset-level
training.
Together, these results show that tabular foundation models can serve not
only as predictors over individual observations, but also as a foundation for
learning reusable representations of supervised datasets.
\clearpage



\section*{Acknowledgments}

We acknowledge the funding support from the "Bayerisches Landesamt für Steuer" for the Bavarian AI Taxation Laboratory. 

Moreover, we gratefully acknowledge the scientific support and HPC resources provided by the Erlangen National High Performance Computing Center (NHR@FAU) of the Friedrich-Alexander-Universität Erlangen-Nürnberg (FAU) under the NHR project DeepSmall - Meta-learning for regularizing deep networks under small data regimes. NHR funding is provided by federal and Bavarian state authorities. NHR@FAU hardware is partially funded by the German Research Foundation (DFG) – 440719683.

\subsection*{Reproducibility statement}

We provide the information needed to reproduce our method and evaluation
throughout the main paper and appendices. Section~\ref{sec:method} specifies
the \methodname{} training objective and architecture-aware dataset readouts,
while Section~\ref{sec:exp} describes the training setup, synthetic
data generation, evaluated models, and primary evaluation protocols.
Appendix~\ref{app:alternative-encoders} provides additional details on the
frozen representations, alternative encoder variants, and metric-learning
objectives. Appendix~\ref{app:results:additional} documents the evaluation
populations, preprocessing, statistical procedures, transformation definitions,
probe protocols, benchmark-selection sensitivity analyses, and task-level
inputs used in the reported experiments. Training and inference code alongside the checkpoints will be released in the future.

\clearpage

\bibliography{references}
\bibliographystyle{iclr2027_conference} 
\appendix

\section{Additional Dataset Representations and \methodname{} Variants}
\label{app:alternative-encoders}

The main text focuses on the pretrained TabICLv2 ICL representation and the
two \methodname{} encoders used in our primary evaluation. Here, we describe
the remaining pretrained representations and learned readouts considered in
our study, together with the alternative metric-learning objectives.

\subsection{Representations from Pretrained TabICLv2}
\label{app:raw-tfm-representations}

TabICLv2 exposes representations at several stages of its
feature-to-row-to-ICL hierarchy
\citep{qu2026tabiclv2betterfasterscalable}: the column, row, and ICL stages
without dataset-level training.

\paragraph{Column mean.}

Let
\begin{equation}
    \mathbf{E}
    \in
    \mathbb{R}^{T\times (G+C_{\mathrm{cls}})\times D}
\end{equation}
denote the output of the pretrained TabICLv2 column embedder, where the
token dimension contains the feature-group and reserved CLS positions. The
column embeddings are target-aware, i.e., the labels $\mathbf{y}$ are
provided to the column embedder.
We obtain a dataset representation by averaging jointly over observations
and token positions,
\begin{equation}
    \representation_{\mathrm{col\text{-}mean}}
    =
    \frac{1}{T(G+C_{\mathrm{cls}})}
    \sum_{t=1}^{T}
    \sum_{m=1}^{G+C_{\mathrm{cls}}}
    \mathbf{E}_{t,m},
    \label{eq:raw-tabicl-col-mean}
\end{equation}
followed by $\ell_2$ normalization. This representation is extracted before
the row-wise and ICL stages of TabICLv2.

\paragraph{Row mean.}

The row interactor processes the column embeddings and produces a
fixed-dimensional representation
\begin{equation}
    \mathbf{r}_t
    \in
    \mathbb{R}^{C_{\mathrm{cls}}D}
\end{equation}
for each observation.
We aggregate these native row representations as
\begin{equation}
    \representation_{\mathrm{row\text{-}mean}}
    =
    \frac{1}{T}
    \sum_{t=1}^{T}
    \mathbf{r}_t.
    \label{eq:raw-tabicl-row-mean}
\end{equation}
The column embedder and row interactor use the pretrained TabICLv2 weights;
no dataset-level adaptation is performed.

\paragraph{ICL mean.}

We next extract the representations produced by TabICLv2's native ICL
transformer. Starting from the row representations, we incorporate the
native target embedding,
\begin{equation}
    \widetilde{\mathbf{r}}_t
    =
    \mathbf{r}_t+\operatorname{YEnc}(y_t),
\end{equation}
for datasets whose class cardinality is supported by the pretrained target
encoder.
The resulting row sequence is processed by the pretrained ICL transformer
with all $T$ observations acting as labeled context,
\begin{equation}
    \mathbf{Q}
    =
    \operatorname{TF}_{\mathrm{ICL}}
    \left(
        [
            \widetilde{\mathbf{r}}_1,\ldots,
            \widetilde{\mathbf{r}}_T
        ];
        \operatorname{train\_size}=T
    \right).
\end{equation}
We apply the native pre-decoder normalization of the ICL predictor and
average the resulting hidden states,
\begin{equation}
    \representation_{\mathrm{ICL\text{-}mean}}
    =
    \frac{1}{T}
    \sum_{t=1}^{T}
    \mathbf{Q}_t,
    \label{eq:raw-tabicl-icl-mean}
\end{equation}
followed by $\ell_2$ normalization. The prediction decoder is not used.
This is the pretrained representation reported as the primary raw-TFM
baseline in the main text.

\paragraph{Randomly initialized control.}

We additionally evaluate whether the representation quality arises from
foundation-model pretraining rather than the architecture and pooling
operation alone. This control uses the same column-embedder and row-interactor
path as the row-mean encoder, but initializes both modules randomly. The
resulting row states are averaged,
\begin{equation}
    \bar{\mathbf{r}}
    =
    \frac{1}{T}
    \sum_{t=1}^{T}\mathbf{r}_t,
\end{equation}
and passed through a randomly initialized projection network,
\begin{equation}
    \embedding_{\mathrm{random}}
    =
    \operatorname{norm}
    \left(
        \projector(\bar{\mathbf{r}})
    \right).
    \label{eq:tabicl-random}
\end{equation}
The ICL transformer is not used in this control. Thus, this baseline
corresponds to a randomly initialized column-to-row encoder with the same
row-mean readout, rather than a randomly initialized full TabICLv2 ICL
stack.

\subsection{Alternative TabICLv2 Encoders}
\label{app:alternative-tabicl}

In addition to the summary-token encoder described in
Sec.~\ref{sec:method:encoders}, we study two alternative mechanisms for
constructing learned dataset representations from TabICLv2
\citep{qu2026tabiclv2betterfasterscalable}.

\paragraph{Contrastively trained row mean.}

The first learned variant retains the native row-mean aggregation,
\begin{equation}
    \representation_{\mathrm{mean}}
    =
    \frac{1}{T}
    \sum_{t=1}^{T}\mathbf{r}_t,
    \label{eq:tabicl-trained-mean}
\end{equation}
but optimizes the TabICLv2 backbone and projection network with the
dataset-level contrastive objective. This provides a direct comparison
between using pretrained row representations as-is and adapting the same
representation to the dataset-level objective while holding the readout
fixed.

\paragraph{Column-aware encoder.}

The native TabICLv2 row representation reduces the feature dimension before
dataset-level aggregation. We instead construct the dataset representation
from the contextual feature states produced immediately before the final
CLS aggregation of the row interactor. Let
\begin{equation}
    \mathbf{F}_{\mathrm{TabICL}}
    \in
    \mathbb{R}^{T\times M\times D}
\end{equation}
denote these feature representations after removing the CLS positions.

We first aggregate each feature representation across observations,
\begin{equation}
    \bar{\mathbf{f}}_m
    =
    \frac{1}{T}
    \sum_{t=1}^{T}
    \mathbf{F}_{\mathrm{TabICL},t,m}.
    \label{eq:tabicl-alt-feature-pool}
\end{equation}
We then apply learned self-attention across the resulting dataset-level
feature representations,
\begin{equation}
    \widetilde{\mathbf{F}}
    =
    \operatorname{TF}_{\mathrm{col}}
    \left(
        [\bar{\mathbf{f}}_1,\ldots,\bar{\mathbf{f}}_M]
    \right),
\end{equation}
and pool over the active feature positions,
\begin{equation}
    \representation_{\mathrm{col}}
    =
    \frac{1}{|\mathcal{M}_{\mathrm{active}}|}
    \sum_{m\in\mathcal{M}_{\mathrm{active}}}
    \widetilde{\mathbf{f}}_m.
    \label{eq:tabicl-alt-column-readout}
\end{equation}
This readout therefore preserves the explicit feature dimension through the
dataset-level attention stage rather than first collapsing each observation
to its CLS representation.

\subsection{Alternative TabPFN-2.5 Encoders}
\label{app:alternative-tabpfn}

TabPFN-2.5 maintains contextual states over observations and feature groups
throughout its transformer backbone \citep{grinsztajn2025tabpfn}. In
addition to the column-aware readout described in
Sec.~\ref{sec:method:encoders}, we study hierarchical summary-token and
pseudo-test representations.

\paragraph{Hierarchical summary-token encoder.}

TabPFN-2.5 does not expose the same fixed-dimensional row representation
followed by a separate ICL transformer as TabICLv2. We therefore learn the
feature-to-row and row-to-dataset reductions explicitly.

Let
\begin{equation}
    \mathbf{F}_{\mathrm{TabPFN}}
    \in
    \mathbb{R}^{T\times M_f\times D}
\end{equation}
denote the contextual feature-group states of the labeled observations.
For each observation $t$, $K=4$ shared learned queries attend to its feature
representations. Their outputs are concatenated and normalized,
\begin{equation}
\mathbf{c}*t
=
\operatorname{LN}
\left(
\operatorname{vec}
\left(
\operatorname{CrossAttn}
\left(
\mathbf{Q}*{1:K},
\mathbf{F}_{\mathrm{TabPFN},t,:}
\right)
\right)
\right)
\in
\mathbb{R}^{KD}.
\label{eq:app-tabpfn-row}
\end{equation}
In the evaluated model, $D=192$ and hence $KD=768$.
The target encoder maps the one-hot class label directly into this same
space using a learned linear projection,
\begin{equation}
\operatorname{YEnc}(y_t)
=
\mathbf{W}*y\operatorname{onehot}(y_t)+\mathbf{b}*y
\in
\mathbb{R}^{KD}.
\end{equation}
We add this target representation to each row representation and append a
learned summary token,
\begin{equation}
\mathbf{C}
=
[
\mathbf{c}*1+\operatorname{YEnc}(y_1),
\ldots,
\mathbf{c}*T+\operatorname{YEnc}(y_T),
\mathbf{s}*{\mathrm{sum}}
].
\end{equation}

A dedicated transformer operates across the resulting row sequence, and the
summary position defines the dataset representation,
\begin{equation}
    \representation_{\mathrm{sum}}
    =
    \operatorname{TF}_{\mathrm{sum}}(\mathbf{C})_{T+1}.
    \label{eq:app-tabpfn-summary}
\end{equation}

\paragraph{Pseudo-test encoder.}

The pseudo-test representation instead uses TabPFN-2.5's native
train--test computation.
We sample $K$ observations without replacement and append copies of their
feature vectors to the dataset as unlabeled test rows,
\begin{equation}
    \mathbf{X}'
    =
    \operatorname{concat}
    \left(
        \mathbf{X},
        \mathbf{X}_{\mathcal{J}}
    \right),
    \qquad
    |\mathcal{J}|=K.
    \label{eq:app-pseudotest-input}
\end{equation}
The original $T$ observations remain the labeled context, while the appended
rows are processed as test observations by the native TabPFN ICL
computation.

Let
\begin{equation}
    \mathbf{g}_1,\ldots,\mathbf{g}_K
    \in
    \mathbb{R}^{D}
\end{equation}
denote the target-column hidden states at the resulting pseudo-test
positions. We concatenate them to form
\begin{equation}
    \representation_{\mathrm{ptest}}
    =
    [
        \mathbf{g}_1;\ldots;\mathbf{g}_K
    ]
    \in
    \mathbb{R}^{KD}.
    \label{eq:app-pseudotest}
\end{equation}
The concatenated representation is subsequently mapped through the
projection network. This construction does not introduce an additional
row- or dataset-level attention module; the pseudo-test states are generated
directly by TabPFN-2.5's pretrained in-context computation.

\subsection{Alternative Metric-Learning Objectives}
\label{app:alternative-objectives}

The main \methodname{} models use NT-Xent \citep{chen2020simpleframeworkcontrastivelearning}.

We also study exponential distance and batch-hard triplet objectives while
retaining the same dataset views and encoder architectures.

Let
\begin{equation}
    \embedding_i^A,\embedding_i^B
    \in
    \mathbb{R}^{d_z}
\end{equation}
denote the $\ell_2$-normalized representations of the two views of dataset
$i$.

\paragraph{Exponential distance.}

We define the cross-view Euclidean distance
\begin{equation}
    \delta_{ij}
    =
    \left\|
        \embedding_i^A-\embedding_j^B
    \right\|_2.
\end{equation}
Pairs with $i=j$ correspond to views of the same dataset, while $i\neq j$
denotes different datasets. The objective is
\begin{equation}
    \mathcal{L}_{\mathrm{exp}}
    =
    \left(
        1-
        \frac{1}{B}
        \sum_{i=1}^{B}
        e^{-\gamma\delta_{ii}}
    \right)
    +
    \frac{1}{B(B-1)}
    \sum_{i\neq j}
    e^{-\gamma\delta_{ij}},
    \label{eq:app-exp-distance}
\end{equation}
where $\gamma$ controls the decay with distance. The first term attracts the
two views of each dataset, while the second penalizes small distances
between representations of different datasets.

\paragraph{Batch-hard triplet loss.}

We additionally evaluate a triplet-margin objective
\citep{Schroff_2015}.
For each anchor
\begin{equation}
    \mathbf{a}_i=\embedding_i^A,
    \qquad
    \mathbf{p}_i=\embedding_i^B,
\end{equation}
we select the closest representation in the opposite view originating from
another dataset,
\begin{equation}
    j^\star
    =
    \arg\min_{j\neq i}
    \left\|
        \embedding_i^A-\embedding_j^B
    \right\|_2,
    \qquad
    \mathbf{n}_i=\embedding_{j^\star}^B.
\end{equation}
The batch-hard triplet objective is
\begin{equation}
    \mathcal{L}_{\mathrm{triplet}}
    =
    \frac{1}{B}
    \sum_{i=1}^{B}
    \max
    \left(
        0,\,
        \left\|
            \mathbf{a}_i-\mathbf{p}_i
        \right\|_2
        -
        \left\|
            \mathbf{a}_i-\mathbf{n}_i
        \right\|_2
        +m
    \right),
    \label{eq:app-triplet}
\end{equation}
where $m$ denotes the triplet margin.

\section{Additional Results and Statistical Comparisons}
\label{app:results:additional}

\subsection{Bootstrap View Discrimination and Variant Selection}
\label{app:results:selection}

We select one variant per backbone using bootstrap-view discrimination AUC on 50 synthetic datasets generated from the TabICLv2 Graph-SCM prior \citep{qu2026tabiclv2betterfasterscalable}.
Table~\ref{tab:results-selection-candidates} compares synthetic
bootstrap-view discrimination across readouts, training objectives,
and initialization schemes.
The selected summary-token and column-aware encoders achieve synthetic AUCs of 0.9999 and 0.9996, respectively.
Several candidates approach saturation; we therefore use this benchmark
for variant selection. The selected summary-token and column-aware encoders
achieve synthetic AUCs of $0.9999$ and $0.9996$, respectively.

\begin{table}[t]
\centering
\caption{Synthetic bootstrap-view discrimination AUC across encoder configurations. Row-mean denotes a trained encoder, distinct from the frozen row-mean baseline. }
\label{tab:results-selection-candidates}
\scriptsize
\setlength{\tabcolsep}{3pt}
\begin{tabular}{llllr}
\toprule
Backbone & Readout & Loss & Initialization & AUC \\
\midrule
TabICL & Summary-token & NT-Xent & Pretrained & 0.9999 \\
TabPFN & Column-aware & NT-Xent & Pretrained & 0.9996 \\
TabICL & Row-mean & NT-Xent & Pretrained & 0.9996 \\
TabPFN & Pseudo-test & NT-Xent & Pretrained & 0.9995 \\
TabICL & Row-mean & Triplet & Pretrained & 0.9994 \\
TabPFN & Pseudo-test & Triplet & Pretrained & 0.9994 \\
TabICL & Summary-token & Triplet & Pretrained & 0.9992 \\
TabICL & Row-mean & NT-Xent & Scratch & 0.9985 \\
TabPFN & Column-aware & Triplet & Pretrained & 0.9981 \\
TabICL & Summary-token & NT-Xent & Scratch & 0.9974 \\
TabICL & Summary-token & Triplet & Scratch & 0.9973 \\
TabICL & Summary-token & Exp. distance & Pretrained & 0.9972 \\
TabICL & Row-mean & Triplet & Scratch & 0.9960 \\
TabICL & Row-mean & Exp. distance & Pretrained & 0.9928 \\
\bottomrule
\end{tabular}
\end{table}

\paragraph{Populations and views.}
The real-data population consists of 191 TabBench datasets.
Synthetic datasets are generated with the TabICLv2 prior using 5--100 features, at most ten classes, 5000--50000 rows, and the prior's unpredictable-dataset filters.
All encoders are evaluated on the same source datasets and the same
four sampled views of each dataset.
For each source dataset, we generate four views by sampling rows with replacement and features without replacement.
View sizes range from 500--5000 rows and 5--50 features, capped at the source dimensions.
We $\ell_2$-normalize the resulting embeddings and compute cosine similarities between all distinct view pairs.
Pairs of views from the same source dataset are positive; pairs from different source datasets are negative.

\paragraph{Hard margins.}
For each source dataset, we define the hard margin as the mean
within-dataset pairwise cosine similarity minus the maximum
cross-dataset cosine similarity involving that dataset.
A positive margin indicates that the average within-dataset similarity
exceeds the similarity of the closest competing cross-dataset pair.

We compare both adapted encoders against nine baselines:
PyMFE, HyTrel, Dataset2Vec, the Neural Statistician,
frozen TabPFN mean-pooling, and four TabICL controls:
ICL-mean, row-mean, column-mean, and random initialization.
On TabBench, median margins are 0.0809 for \methodname{}-ICL
and 0.0587 for \methodname{}-PFN, compared with at most 0.0045
among the baselines.
On synthetic datasets, the corresponding medians are 0.5945
and 0.5605, compared with at most 0.0306 among the baselines.

Both adapted encoders significantly outperform every baseline
in each population under two-sided Wilcoxon signed-rank tests
paired by dataset, with Holm correction across all 18 comparisons
within each population.
Table~\ref{tab:results-bootstrap-significance} summarizes these tests.
These comparisons assess margins rather than AUC; the synthetic
results characterize the population used for model selection.

\begin{table}[t]
\centering
\caption{Paired comparisons of per-dataset hard margins in
bootstrap-view discrimination. Both adapted encoders improve over
each of nine baselines in each population. Holm correction is
applied across all 18 comparisons within each population.
The final column gives an upper bound on the largest adjusted
$p$-value. These tests assess margins, not AUC.
Synthetic results characterize the population used for model selection.}
\label{tab:results-bootstrap-significance}
\small
\setlength{\tabcolsep}{4pt}
\begin{tabular}{lrrr}
\toprule
Population & Datasets & Significant improvements & Largest Holm $p$ \\
\midrule
TabBench & 191 & 18/18 & $<1.6\times10^{-16}$ \\
Synthetic & 50 & 18/18 & $<5.0\times10^{-14}$ \\
\bottomrule
\end{tabular}
\end{table}

\subsection{Retrieval Robustness under Dataset Transformations}
\label{app:results:transformations}
For paired inference, we average recall@5 across applicable
transformations within each dataset.
Two-sided Wilcoxon signed-rank tests compare these dataset-level
scores on the population shared by each encoder pair, dropping
zero differences.
The four comparisons of the two adapted encoders against
TabICL ICL-mean and PyMFE form a single Holm-correction family
(Table~\ref{tab:results-significance}).
Both adapted encoders significantly outperform TabICL ICL-mean,
while \methodname{}-PFN also significantly outperforms PyMFE.
These tests concern dataset-level averages, whereas the headline
retrieval score weights the 19 transformations equally.

\begin{table}[t]
\centering
\caption{Dataset transformations used for retrieval evaluation.
Here, $n$ and $d$ denote the original numbers of observations and
features, and $\mu_j$ and $\sigma_j$ denote the mean and standard
deviation of numerical feature $j$.
We write
$k_r=\max\{\min(n,20),\operatorname{round}(rn)\}$ and
$h_r=\max\{1,\operatorname{round}(rd)\}$.
Grouped entries denote separate transformations for each listed
parameter value.}
\label{tab:results-transform-protocol}
\scriptsize
\setlength{\tabcolsep}{4pt}
\renewcommand{\arraystretch}{1.15}
\begin{tabular}{p{0.25\linewidth}p{0.68\linewidth}}
\toprule
Transformation & Definition \\
\midrule

Keep 20\% / 50\% of rows &
Sample $k_r$ observations without replacement,
$r\in\{0.2,0.5\}$. Skip when $k_r\geq n$. \\

Bootstrap rows &
Sample $n$ observations with replacement. \\

Duplicate rows &
Retain all observations and append
$\max\{1,\operatorname{round}(0.3n)\}$ observations
sampled with replacement. \\

Keep 25\% / 50\% / 75\% of features &
Sample $h_r$ features without replacement,
$r\in\{0.25,0.5,0.75\}$. Skip when $h_r\geq d$. \\

Duplicate features &
Append copies of $h_{0.3}$ features selected without replacement. \\

Add noisy features &
Append $h_{0.3}$ numerical features with independent
$\mathcal{N}(0,1)$ entries. \\

Light / heavy noise &
Add independent $\mathcal{N}(0,(\alpha\sigma_j)^2)$ noise
to numerical features, with $\alpha\in\{0.1,0.5\}$.
Features with zero or non-finite standard deviation are skipped. \\

Affine rescaling &
For each numerical feature, apply $x'=a_jx+b_j$, where
$\log a_j\sim\mathcal{U}(\log 0.2,\log 5)$ and
$b_j\sim\mathcal{N}(0,\sigma_j^2)$. \\

Monotone power &
For each numerical feature, set $t=x/\sigma_j$ and apply
$x'=\operatorname{sign}(t)|t|^{a_j}$,
where $a_j\sim\mathcal{U}(0.5,2)$. \\

Quantization &
For numerical features with more than ten distinct observed
values, assign integer codes using ten quantile bins of
observation ranks. Ties are broken by observation order;
existing missing values are preserved. \\

Standardization &
Apply $x'=(x-\mu_j)/\sigma_j$ to numerical features. \\

10\% / 30\% missingness &
Independently replace each feature entry with a missing value
with probability $0.1$ or $0.3$, including categorical entries. \\

Categorical recoding &
Bijectively relabel the observed categories using a random
permutation, for each non-numerical feature with at least
two distinct observed values. \\

Combined view &
Sample $k_{0.7}$ observations with replacement and $h_{0.7}$
features without replacement, then apply light numerical noise
($\alpha=0.1$). Retain the sampled view if numerical noise
is inapplicable. \\

\bottomrule
\end{tabular}
\end{table}

\begin{figure}[t]
\centering
\IfFileExists{results_figures/results_retrieval_full_plot.pdf}{\includegraphics[width=\linewidth]{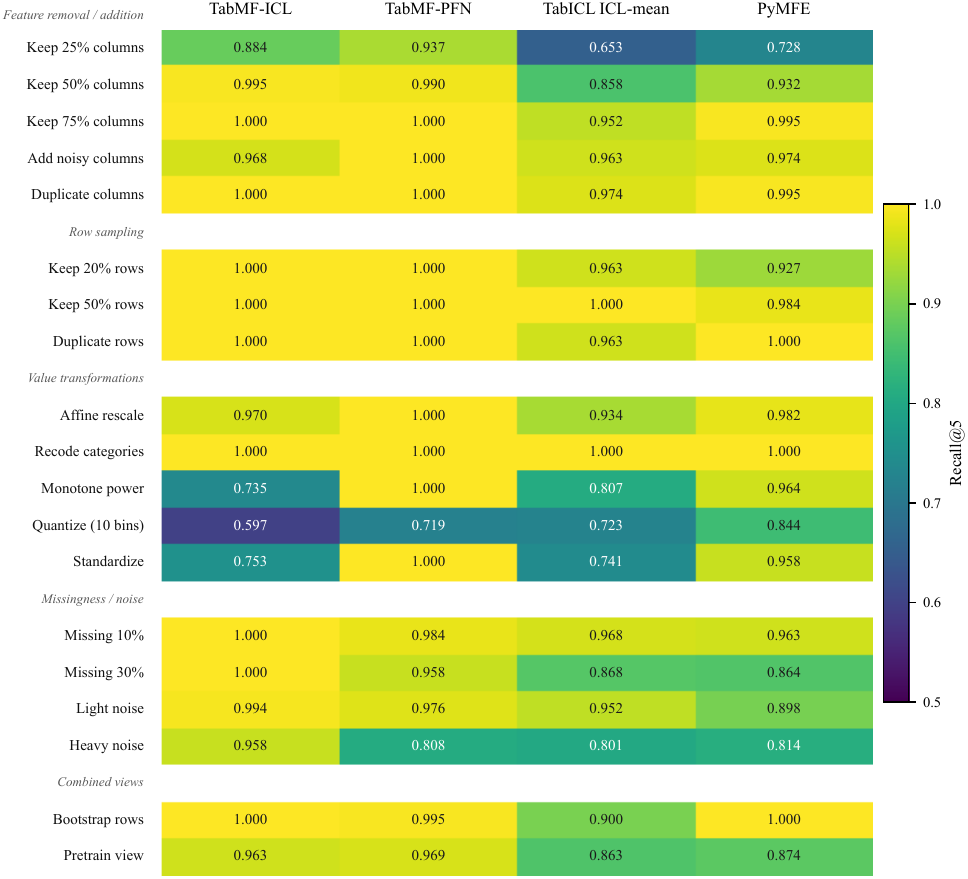}}{\fbox{\parbox[c][65mm][c]{0.94\linewidth}{\centering Complete retrieval heatmap: 19 transformations and four encoders.}}}
\caption{Retrieval recall@5 under all 19 dataset transformations
for \methodname{}-ICL, \methodname{}-PFN, frozen TabICL ICL-mean,
and PyMFE. The headline retrieval score averages recall@5 equally
across these transformations.
Dataset-level paired comparisons are reported in
Table~\ref{tab:results-significance}.}
\label{fig:results-transforms-full}
\end{figure}

Figure~\ref{fig:results-transforms-full} reports recall@5 for all
19 transformations.

\begin{table}[t]
\centering
\small
\caption{Dataset-level paired comparisons of transformation retrieval.
For each dataset, recall@5 is averaged across applicable transformations.
Entries are Holm-adjusted $p$-values from two-sided Wilcoxon
signed-rank tests, correcting across all four comparisons.
Tests pair 190 datasets, except \methodname{}-PFN versus PyMFE,
which pairs 191.
Bold denotes a significant improvement of the adapted encoder.
These tests assess dataset-level averages; the headline retrieval
score instead averages recall@5 equally across transformations.}
\label{tab:results-significance}
\begin{tabular}{lrr}
\toprule
Baseline & \methodname{}-ICL & \methodname{}-PFN \\
\midrule
TabICL ICL-mean & $\mathbf{1.0\times10^{-7}}$
               & $\mathbf{5.8\times10^{-15}}$ \\
PyMFE          & $0.522$
               & $\mathbf{1.6\times10^{-5}}$ \\
\bottomrule
\end{tabular}
\end{table}

\subsection{Three-Dataset Benchmark Selection}
\label{app:results:benchmark-selection}

We apply the benchmark-selection framework of
\citet{gusev2026benchmarking} to TabRepo
\citep{salinas2024tabrepolargescalerepository}, using 144
classification datasets and ten default classifiers.
Each comparison uses 200 shared candidate pools containing
80\% of the population and the same full-population classifier
ranking as reference.
Embeddings are standardized and $\ell_2$-normalized before either
K-means representative selection or cosine farthest-first selection.
Each rule selects three datasets.
Table~\ref{tab:results-benchmark-selection} reports the results
for both selection rules.
Only K-means selection contributes to the primary aggregate rank;
farthest-first selection provides an additional evaluation.

\begin{table}[t]
\centering
\caption{
Compact benchmark selection on TabRepo at $k=3$.
Entries are mean classifier-ranking Spearman correlations over 200 shared
candidate-pool trials.
Encoders are ordered by K-means performance.
Bold marks numerical leaders.
}
\label{tab:results-benchmark-selection}
\begin{tabular}{lcc}
\toprule
Encoder & K-means & Farthest-first \\
\midrule
\methodname{}-PFN      & \textbf{0.9292} & \textbf{0.8229} \\
\methodname{}-ICL      & 0.8542 & 0.8146 \\
TabPFN mean-pooling    & 0.8500 & 0.6730 \\
Random                 & 0.7766 & 0.7766 \\
Neural Statistician & 0.738 & 0.782 \\
PyMFE                  & 0.7093 & 0.7812 \\
Dataset2Vec            & 0.6515 & 0.7261 \\
TabICL ICL-mean        & 0.5750 & 0.7108 \\
\bottomrule
\end{tabular}
\end{table}

\paragraph{Budget sensitivity and statistical tests.}
We use $k=3$ as the primary compact-benchmark setting and report
sensitivity over $k\in\{2,3,4\}$ below. Because $k=3$ was selected
during the evaluation analysis, we interpret the corresponding significance
tests together with this budget-sensitivity analysis.
Paired Wilcoxon signed-rank tests use the 200 shared candidate-pool trials.
For each selection rule, the comparisons of the two adapted encoders
against Random, TabICL ICL-mean, PyMFE, Dataset2Vec, the Neural
Statistician, and frozen TabPFN mean-pooling form a single
12-comparison Holm-correction family.
Table~\ref{tab:results-benchmark-tests} reports the corresponding
Holm-adjusted $p$-values for K-means.
The reported 95\% confidence intervals are percentile-bootstrap
intervals for mean paired differences across candidate-pool trials.
\begin{table}[t]
\centering
\small
\caption{K-means benchmark-selection comparisons on TabRepo
at a three-dataset budget. Differences are \methodname{} minus
baseline in mean Spearman correlation.
Two-sided Wilcoxon signed-rank tests pair the 200 candidate-pool trials.
Reported $p$-values are Holm-adjusted jointly over the 12 comparisons.}
\label{tab:results-benchmark-tests}
\setlength{\tabcolsep}{5pt}
\begin{tabular}{llrr}
\toprule
Variant & Baseline & $\Delta\rho$ & Holm $p$ \tabularnewline
\midrule
\methodname{}-ICL & Random
& $+0.0777$ & $5.28\times10^{-7}$ \tabularnewline
\methodname{}-ICL & ICL-mean
& $+0.2792$ & $6.03\times10^{-28}$ \tabularnewline
\methodname{}-ICL & PyMFE
& $+0.1450$ & $2.67\times10^{-22}$ \tabularnewline
\methodname{}-ICL & Dataset2Vec
& $+0.2028$ & $1.74\times10^{-25}$ \tabularnewline
\methodname{}-ICL & Neural Statistician
& $+0.1162$ & $2.00\times10^{-19}$ \tabularnewline
\methodname{}-ICL & Frozen TabPFN
& $+0.0045$ & $0.2408$ \tabularnewline
\midrule
\methodname{}-PFN & Random
& $+0.1526$ & $2.18\times10^{-23}$ \tabularnewline
\methodname{}-PFN & ICL-mean
& $+0.3542$ & $2.07\times10^{-32}$ \tabularnewline
\methodname{}-PFN & PyMFE
& $+0.2199$ & $7.27\times10^{-32}$ \tabularnewline
\methodname{}-PFN & Dataset2Vec
& $+0.2777$ & $2.51\times10^{-32}$ \tabularnewline
\methodname{}-PFN & Neural Statistician
& $+0.1911$ & $2.32\times10^{-31}$ \tabularnewline
\methodname{}-PFN & Frozen TabPFN
& $+0.0795$ & $2.03\times10^{-24}$ \tabularnewline
\bottomrule
\end{tabular}
\end{table}

\paragraph{Frozen TabPFN comparison.}
On TabRepo, the mean paired improvements of \methodname{}-PFN over frozen
TabPFN mean-pooling are 0.0795 with K-means
(95\% bootstrap interval $[0.068,0.091]$) and 0.1500 with farthest-first
selection ($[0.133,0.167]$).
Under the 12-comparison Holm family defined separately for each selector,
the corresponding adjusted Wilcoxon $p$-values are
$2.03\times10^{-24}$ and $1.96\times10^{-26}$.
\methodname{}-ICL does not significantly exceed frozen TabPFN under K-means
(Holm-adjusted $p=0.2408$), but does under farthest-first selection
($p=1.62\times10^{-25}$).
All of these encoders have complete coverage of the evaluated TabRepo population.

\paragraph{Robustness checks.}
An alternative K-means initialization preserves the direction
of all six comparisons between the adapted encoders and
TabICL ICL-mean, PyMFE, and Dataset2Vec
(Table~\ref{tab:results-benchmark-seeds}).
Across $k\in\{2,3,4\}$, a \methodname{} variant obtains the highest
numerical score under both K-means and farthest-first selection.
The strongest variant depends on the selection budget:
\methodname{}-ICL leads K-means at $k=2$, whereas
\methodname{}-PFN leads at $k=3$ and $k=4$ and under
farthest-first selection at all three budgets
(Table~\ref{tab:results-benchmark-budgets}).

\begin{table}[t]
\centering
\small
\caption{Sensitivity of three-dataset selection on TabRepo
to K-means initialization.
Entries are paired mean-correlation differences
(\methodname{} minus baseline).
Candidate pools and embeddings are fixed.
The alternative initialization preserves the direction
of all six comparisons.}
\label{tab:results-benchmark-seeds}
\setlength{\tabcolsep}{5pt}
\begin{tabular}{llrr}
\toprule
Variant & Baseline & Original & Alternative \\
\midrule
\methodname{}-ICL & ICL-mean    & $+0.2792$ & $+0.2602$ \\
\methodname{}-ICL & PyMFE       & $+0.1450$ & $+0.1280$ \\
\methodname{}-ICL & Dataset2Vec & $+0.2028$ & $+0.1946$ \\
\methodname{}-PFN & ICL-mean    & $+0.3542$ & $+0.3478$ \\
\methodname{}-PFN & PyMFE       & $+0.2199$ & $+0.2155$ \\
\methodname{}-PFN & Dataset2Vec & $+0.2777$ & $+0.2822$ \\
\bottomrule
\end{tabular}
\end{table}
\begin{table}[t]
\centering
\caption{TabRepo benchmark selection with two to four selected datasets.
Entries are mean Spearman correlations with the full-population classifier
ranking. Bold marks the best score within each fixed budget and selector.
Various budgets assess sensitivity to the selection budget.}
\label{tab:results-benchmark-budgets}
\scriptsize
\setlength{\tabcolsep}{2pt}

\begin{tabular}{rlrrrrrrrr}
\toprule
$k$ & Selector & \methodname{}-ICL & \methodname{}-PFN & ICL-mean & PyMFE &
Random & D2V & Frozen PFN & Neural Stat. \tabularnewline
\midrule
2 & K-means & \textbf{0.8568} & 0.6728 & 0.3507 & 0.6778 &
0.7209 & 0.6947 & 0.8214 & 0.7828 \tabularnewline

2 & Farthest-first & 0.7003 & \textbf{0.8752} & 0.5941 & 0.6979 &
0.7209 & 0.7100 & 0.6439 & 0.6418 \tabularnewline

3 & K-means & 0.8542 & \textbf{0.9292} & 0.5750 & 0.7093 &
0.7766 & 0.6515 & 0.8497 & 0.7380 \tabularnewline

3 & Farthest-first & 0.8146 & \textbf{0.8229} & 0.7108 & 0.7812 &
0.7766 & 0.7261 & 0.6728 & 0.7819 \tabularnewline

4 & K-means & 0.8504 & \textbf{0.9393} & 0.6743 & 0.8894 &
0.8171 & 0.7282 & 0.8536 & 0.7260 \tabularnewline

4 & Farthest-first & 0.8127 & \textbf{0.8697} & 0.8148 & 0.8338 &
0.8171 & 0.7805 & 0.7447 & 0.8560 \tabularnewline
\bottomrule
\end{tabular}
\end{table}

\paragraph{Coverage.}
HyTrel embeds 129 of the 144 TabRepo datasets.
Every shared candidate pool contains at least one dataset
without a HyTrel embedding, so HyTrel has no score under
the common-population protocol.
The Neural Statistician covers all 144 datasets;
its results under both selection rules are reported in
Table~\ref{tab:results-benchmark-selection}.

\subsection{Dataset-Property Probes}
\label{app:results:properties}

\subsubsection{Probe Protocol}

Encoders receive complete tables and their designated predictive targets when supported; regression targets are quantile-binned into at most ten classes where required by the encoder.
Property labels and scores are used only by the downstream probes.
We use tuned linear probes for the primary evaluation because they provide a common and interpretable measure of how readily a property can be recovered from each representation.

For these probes, robust median/IQR scaling is followed by clipping to $\pm8$ and PCA retaining 99.9\% of variance.
Preprocessing is refitted within the inner folds of a grid search selecting among 17 logistic-regression penalties or 40 ridge penalties.
Classification uses stratified folds and log-loss model selection; regression uses ordinary folds and an SVD ridge solver.
The three-fold outer evaluation is repeated five times.
PyMFE missing values are median-imputed over the evaluation population before cross-validation; the other checked embeddings contain no missing values.

Domain and paired property analyses use a common population of 190 datasets for the main encoder comparisons.
Available-case regression coverage is 190 or 191 depending on the encoder, while HyTrel property probes cover 177 datasets.
The primary aggregate includes encoders with reported scores on all six ranking units; individual probe analyses use their stated available-case populations.
\subsubsection{Domain Discrimination}
\label{app:results:tabbench}

The TabBench datasets are assigned to eight broad domains: Healthcare, Behavioral, Computer Vision, Industry \& Science, Other, Finance/insurance, Games \& Synthetic, and Social/Public.
The primary endpoint uses the tuned logistic-regression probe above and averages fold-level macro one-vs-rest AUCs within repetitions.

A separate paired dataset-bootstrap analysis uses the common 190-dataset population and the mean of five pooled-out-of-fold AUCs.
Its point estimates are 0.697 for \methodname{}-ICL, 0.681 for \methodname{}-PFN, 0.661 for TabICL ICL-mean, 0.653 for TabPFN mean-pooling, and 0.641 for Dataset2Vec.
None of the ten pairwise comparisons remains significant after Holm correction.
This analysis applies to the pooled-out-of-fold endpoint and is therefore not attached to the fold-averaged domain scores in Table~\ref{tab:results-primary}.

\paragraph{Strengthened nonlinear probe.}
To assess whether the linear comparison is sensitive to readout capacity, we repeat domain discrimination with a tuned SVM that jointly searches RBF and polynomial kernels after the same fold-local preprocessing.
Table~\ref{tab:results-industry-strengthened-probe} reports all 11 encoders in the restricted comparison pool.
Each encoder's summary metric uses its maximal available population; pairwise tests use the population shared by the corresponding pair.
Significance is evaluated with paired Wilcoxon tests on per-dataset log-loss and Holm correction over all 55 encoder pairs.
Under this stronger probe, \methodname{}-ICL outperforms frozen TabICL ICL-mean (AUC 0.6997 versus 0.6441; adjusted $p=0.0278$), whereas \methodname{}-PFN is not significantly separated from frozen TabPFN mean-pooling.

\begin{table}[t]
\centering
\caption{Domain discrimination with the strengthened nonlinear probe. A tuned SVM jointly searches RBF and polynomial kernels. AUC and log-loss are computed on each encoder's maximal available population. Pairwise inference uses per-dataset log-loss on the population shared by each pair, with Holm correction over all 55 encoder comparisons. Bold denotes the best point estimate.}
\label{tab:results-industry-strengthened-probe}
\small
\begin{tabular}{lrrr}
\toprule
Encoder & AUC $\uparrow$ & Log-loss $\downarrow$ & $N$ \\
\midrule
\methodname{}-ICL              & \textbf{0.6997} & \textbf{1.7860} & 190 \\
\methodname{}-PFN              & 0.6808          & 1.8100          & 191 \\
PyMFE                           & 0.6662          & 1.8871          & 191 \\
TabICL row-mean                 & 0.6637          & 1.8850          & 190 \\
TabPFN mean-pooling             & 0.6620          & 1.8535          & 191 \\
TabICL ICL-mean                 & 0.6441          & 1.8585          & 190 \\
Dataset2Vec                     & 0.6239          & 1.9286          & 191 \\
HyTrel                          & 0.6158          & 1.9798          & 177 \\
Neural Statistician             & 0.5517          & 2.0017          & 191 \\
TabICL column-mean              & 0.5442          & 2.0313          & 190 \\
TabICL random initialization    & 0.5256          & 2.0322          & 190 \\
\bottomrule
\end{tabular}
\end{table}

\subsubsection{Epiplexity Prediction}
\label{app:results:epiplexity}

We assign each dataset a prequential epiplexity score motivated by learnable information for computationally bounded observers \citep{finzi2026entropy}.
The estimate is specific to the observer and protocol described here and should not be interpreted as model-independent prediction difficulty.
Categorical variables are integer-coded, missing values are median-imputed, and features are z-scored.
Rows are shuffled once with seed 42 and processed in blocks of 32.
An online SGD classifier with logistic loss, an adaptive learning-rate schedule, and initial learning rate $\eta_0=0.01$ predicts each block before updating on it.
The epiplexity meter uses baseline entropy $\ln C$, corresponding to a uniform distribution over the $C$ classes, and returns the final dataset-level score.

\paragraph{Primary linear-probe inference.}
For paired comparisons, five encoders are evaluated on the same 190 datasets, ordered by dataset identifier and using shared cross-validation partitions.
Within each of five repetitions, Spearman correlation is computed over pooled out-of-fold predictions from the tuned ridge probe.
We jointly resample dataset identifiers in 10,000 paired bootstrap draws; correlations are computed separately for each repetition and averaged within each draw.
Holm correction is applied across all ten pairwise encoder comparisons.
These tests apply to Table~\ref{tab:results-epiplexity-common}, whereas Table~\ref{tab:results-primary} reports available-case scores using each encoder's native coverage.
Cross-validation repetitions measure sensitivity to the fold partition and are not treated as independent datasets.

\begin{table}[t]
\centering
\small
\setlength{\tabcolsep}{5pt}
\caption{Epiplexity prediction on a common population of 190 datasets. \textbf{A:} Mean Spearman correlation across five cross-validation repetitions. \textbf{B:} Frozen-versus-adapted comparisons from a paired dataset bootstrap, with Holm adjustment over all ten pairwise comparisons among the five encoders. Bold $p$-values support higher correlation for the frozen readout.}
\label{tab:results-epiplexity-common}
\textbf{A. Common-population scores}\par\smallskip
\begin{tabular}{lr}
\toprule
Encoder & Spearman $\rho$ \\
\midrule
TabICL ICL-mean & 0.764 \\
TabPFN mean-pooling & 0.760 \\
\methodname{}-ICL & 0.639 \\
\methodname{}-PFN & 0.621 \\
Dataset2Vec & 0.529 \\
\bottomrule
\end{tabular}
\par\medskip
\textbf{B. Frozen-versus-adapted comparisons}\par\smallskip
\begin{tabular}{llr}
\toprule
Frozen readout & Adapted encoder & Holm $p$ \\
\midrule
TabICL ICL-mean & \methodname{}-ICL & \textbf{0.040} \\
TabICL ICL-mean & \methodname{}-PFN & \textbf{0.040} \\
TabPFN mean-pooling & \methodname{}-ICL & 0.079 \\
TabPFN mean-pooling & \methodname{}-PFN & \textbf{0.040} \\
\bottomrule
\end{tabular}
\end{table}

On the common population, TabICL ICL-mean significantly outperforms both adapted encoders, and TabPFN mean-pooling significantly outperforms \methodname{}-PFN.
The remaining frozen-versus-adapted comparison is not significant after correction.
The two frozen encoders are not significantly separated from one another, nor are the two adapted encoders.

\paragraph{Strengthened nonlinear probe.}
We repeat epiplexity prediction with kernel ridge regression and a substantially wider joint search over RBF, Laplacian, and polynomial kernels after the same fold-local preprocessing.
Table~\ref{tab:results-epiplexity-strengthened-probe} reports the full 11-encoder restricted pool, using the leakage-controlled PyMFE representation.
The primary metric for this capacity check is mean per-dataset absolute error, which is also the unit used by the paired Wilcoxon tests; Spearman correlation is reported for comparability with the primary evaluation.
As above, summary metrics use each encoder's maximal population and pairwise tests use the population shared by each pair, with Holm correction over all 55 comparisons.
Under this stronger readout, neither \methodname{} variant is significantly separated from its corresponding frozen representation.
Thus, the primary ridge results establish differences in linear accessibility, but do not establish that epiplexity-related information is absent from the adapted embeddings.

\begin{table}[t]
\centering
\caption{Epiplexity prediction with the strengthened nonlinear probe. Kernel ridge regression jointly searches RBF, Laplacian, and polynomial kernels. Mean absolute error is the endpoint used for ranking and paired inference; Spearman correlation is included for comparison with the primary linear probe. Pairwise tests use the population shared by each pair, with Holm correction over all 55 comparisons. Bold denotes the best point estimate.}
\label{tab:results-epiplexity-strengthened-probe}
\small
\begin{tabular}{lrrr}
\toprule
Encoder & Mean $|\mathrm{error}|$ $\downarrow$ & Spearman $\uparrow$ & $N$ \\
\midrule
TabPFN mean-pooling             & \textbf{0.1770} & \textbf{0.8292 $\pm$ 0.026} & 191 \\
\methodname{}-PFN              & 0.1794          & 0.7633 $\pm$ 0.019          & 191 \\
TabICL ICL-mean                 & 0.2032          & 0.7991 $\pm$ 0.021          & 190 \\
\methodname{}-ICL              & 0.2224          & 0.6953 $\pm$ 0.012          & 190 \\
Dataset2Vec                     & 0.2265          & 0.6182 $\pm$ 0.026          & 191 \\
Neural Statistician             & 0.3297          & 0.2658 $\pm$ 0.027          & 191 \\
HyTrel                          & 0.3348          & 0.1746 $\pm$ 0.038          & 177 \\
TabICL column-mean              & 0.3447          & 0.2501 $\pm$ 0.043          & 190 \\
TabICL row-mean                 & 0.3465          & 0.2289 $\pm$ 0.092          & 190 \\
TabICL random initialization    & 0.3511          & 0.1898 $\pm$ 0.052          & 190 \\
PyMFE (leakage-controlled)      & 0.3703          & 0.1160 $\pm$ 0.036          & 191 \\
\bottomrule
\end{tabular}
\end{table}

\subsubsection{PyMFE Control for Epiplexity}
\label{app:results:pymfe}

Full PyMFE includes general, statistical, information-theoretic, model-based, and landmarking descriptors fitted to $(\mathbf{X},\mathbf{y})$.
For epiplexity prediction, we instead use only the general, statistical, and information-theoretic groups, fit the extractor without $\mathbf{y}$, omit the model-based and landmarking groups, and remove all summaries of the target-dependent base features listed in Table~\ref{tab:results-pymfe-exclusions}.
This controls direct overlap with ingredients of the epiplexity score, including its class-count-dependent baseline.
Full PyMFE is retained for the other evaluation tasks.

\begin{table}[t]
\centering
\caption{Explicit exclusions in the leakage-controlled PyMFE difficulty configuration. Landmarking and model-based groups are omitted entirely. The remaining general, statistical, and information-theoretic groups are fitted without the dataset target. All summaries of each listed base feature are excluded together.}
\label{tab:results-pymfe-exclusions}
\small
\begin{tabular}{lll}
\toprule
\multicolumn{3}{c}{Excluded base-feature identifiers} \\
\midrule
\texttt{nr\_class} & \texttt{freq\_class} & \texttt{class\_ent} \\
\texttt{class\_conc} & \texttt{mut\_inf} & \texttt{joint\_ent} \\
\texttt{eq\_num\_attr} & \texttt{ns\_ratio} & \texttt{can\_cor} \\
\texttt{gravity} & \texttt{lh\_trace} & \texttt{p\_trace} \\
\texttt{roy\_root} & \texttt{w\_lambda} & \texttt{sd\_ratio} \\
\bottomrule
\end{tabular}
\end{table}

\subsection{Aggregate Ranking and Statistical Scope}
\label{app:results:statistics}

The primary aggregate combines six evaluation units: real-data bootstrap-view discrimination AUC, domain macro-OVR AUC, continuous epiplexity Spearman correlation, mean transformation-retrieval recall@5, TabRepo three-dataset K-means ranking Spearman, and TableShift shift-severity Spearman correlation.
Each unit ranks the same seven encoders with complete task coverage, using average ranks for ties.
Synthetic model-selection scores, farthest-first selection,
web-table transfer, and hyperparameter transfer do not enter
the aggregate.
No significance test is applied to the aggregate ranks. 

We assess sensitivity to the benchmark-selection task by varying
the K-means selection budget over $k\in\{2,3,4\}$ and by omitting
the task entirely.
All comparisons retain the same seven encoders and leave the
other five evaluation units unchanged
(Table~\ref{tab:results-aggregate-sensitivity}).
Both \methodname{} variants retain lower median ranks than every
baseline in all four settings.
Their relative ordering depends on the selection budget:
\methodname{}-ICL leads at $k=2$, whereas \methodname{}-PFN leads
at $k=3$ and $k=4$.
Without benchmark selection, both obtain a median rank of 2.0.

\begin{table}[t]
\centering
\small
\caption{Sensitivity of the six-task aggregate ranking to the TabRepo
benchmark-selection budget. Only the TabRepo K-means evaluation changes
across columns; the other five evaluation units are held fixed.
The aggregate score is the median rank across the six evaluation units.}
\label{tab:results-aggregate-sensitivity}
\setlength{\tabcolsep}{4pt}

\begin{tabular}{lrrrrrr}
\toprule
Encoder &
TR $k=2$ &
TR $k=3$ &
TR $k=4$ &
Agg. $k=2$ &
Agg. $k=3$ &
Agg. $k=4$
\tabularnewline
\midrule

\methodname{}-PFN
& 6 & 1 & 1
& 2.0 & \textbf{1.5} & \textbf{1.5}
\tabularnewline

\methodname{}-ICL
& 1 & 2 & 4
& \textbf{1.5} & 2.0 & 2.0
\tabularnewline

TabICL ICL-mean
& 7 & 7 & 7
& 3.5 & 3.5 & 3.5
\tabularnewline

PyMFE
& 5 & 5 & 2
& 4.0 & 4.0 & 3.5
\tabularnewline

Frozen TabPFN
& 2 & 3 & 3
& 5.0 & 5.0 & 5.0
\tabularnewline

Dataset2Vec
& 4 & 6 & 5
& 5.0 & 5.5 & 5.0
\tabularnewline

Neural Statistician
& 3 & 4 & 6
& 7.0 & 7.0 & 7.0
\tabularnewline

\bottomrule
\end{tabular}
\end{table}

Bootstrap hard-margin and benchmark-selection tests use the pairing units and correction families specified in their respective subsections.
Probe bootstraps resample dataset identifiers while retaining the repeated predictions associated with each sampled identifier.
They characterize variation over the evaluated dataset population conditional on the fitted predictions and do not include retraining within each bootstrap draw.
TableShift contains nine task-level observations; we therefore report the observed correlations without asserting pairwise encoder superiority.

\paragraph{TableShift shift-severity inputs.}
Table~\ref{tab:tableshift-inputs} reports the nine task-level inputs used for
the shift-severity evaluation. For each task, the severity target is the mean
ID accuracy minus the mean OOD accuracy over the TableShift estimators for
which both published values are available. Thus, each task contributes one
severity value to the Spearman correlation; no estimator-level observations
are treated as independent samples.

\begin{table}[t]
\centering
\small
\caption{TableShift tasks used for the shift-severity evaluation.
The accuracy gap is the mean ID-minus-OOD accuracy difference across
estimators with both published ID and OOD results. Row counts are the
actual ID and OOD samples embedded after the at-most-8,192-row sampling
described in Section~\ref{sec:exp}.}
\label{tab:tableshift-inputs}
\begin{tabular}{llrrr}
\toprule
Task & Dataset ID & Gap & ID rows & OOD rows \\
\midrule
ASSISTments          & \texttt{assistments}             & 0.3519 & 8192 & 1906 \\
FICO HELOC           & \texttt{heloc}                   & 0.2826 & 278  & 6914 \\
College Scorecard    & \texttt{college\_scorecard}      & 0.1132 & 8192 & 1352 \\
Hypertension         & \texttt{brfss\_blood\_pressure}  & 0.0816 & 8192 & 8192 \\
Hospital Readmission & \texttt{diabetes\_readmission}   & 0.0489 & 4287 & 8192 \\
Diabetes             & \texttt{brfss\_diabetes}         & 0.0450 & 8192 & 8192 \\
Food Stamps          & \texttt{acsfoodstamps}           & 0.0261 & 8192 & 8192 \\
Income               & \texttt{acsincome}               & 0.0221 & 8192 & 8192 \\
Unemployment         & \texttt{acsunemployment}         & 0.0193 & 8192 & 8192 \\
\bottomrule
\end{tabular}
\end{table}

\subsection{Encoder Variants}
\label{app:results:ablations}

Table~\ref{tab:results-ablations} compares retrieval performance across
initialization schemes, training objectives, and readouts. For the
initialization comparison, architecture, readout, objective, and batch regime
are held fixed within each backbone. Pretrained initialization outperforms
scratch initialization for both TabICL and TabPFN.

For the remaining TabPFN variants, the pretrained column-aware encoder with
NT-Xent obtains $0.965$ recall, compared with $0.935$ for the summary-token
encoder and $0.957$ for the pseudo-test encoder.
\subsection{Web-Table and Hyperparameter Transfer}
\label{app:results:transfer}

Table~\ref{tab:results-transfer} reports web-table and hyperparameter-transfer evaluations separately from the primary suite.

For web-table transfer, TEmBed similarity averages MAP over 16 GitTables datalakes \citep{hulsebos2023gittables} for the TFM encoders and HyTrel, while the PyMFE aggregate covers 17 datalakes and is marked accordingly.
Shuffling covers 60 slices.
Reported standard deviations measure variation across datalakes or slices rather than confidence intervals.
Similarity and shuffling use uncapped table inputs.
These evaluations concern target-agnostic table representations and
therefore differ from the supervised-dataset representation setting
considered in the main evaluation.

For hyperparameter transfer, aggregates cover 82 targets across 16 search
spaces for the adapted encoders and TabICL ICL-mean, compared with 77 targets
for HyTrel. Within each search space, embeddings are median-imputed and
standardized, and each target retrieves its single Euclidean-nearest
meta-training dataset ($m=1$). We transfer the five configurations ranked
highest on that source dataset and report the minimum normalized target regret;
neighbor quality normalizes the corresponding transfer performance between a
random-source reference (zero) and an oracle-source reference (one).
Aggregates are pooled across spaces by target count. 
\subsection{Encoder Efficiency and Scaling}
\label{app:results:efficiency}

We additionally characterize the computational cost of producing dataset
representations as table size increases. We vary the number of observations from
500 to 200,000 while fixing the number of features to 64, and vary the number
of features from 8 to 2,048 while fixing the number of observations to 10,000.
Each setting is evaluated on three independently generated synthetic datasets.

\begin{figure}[t]
    \centering
    \includegraphics[width=\linewidth]{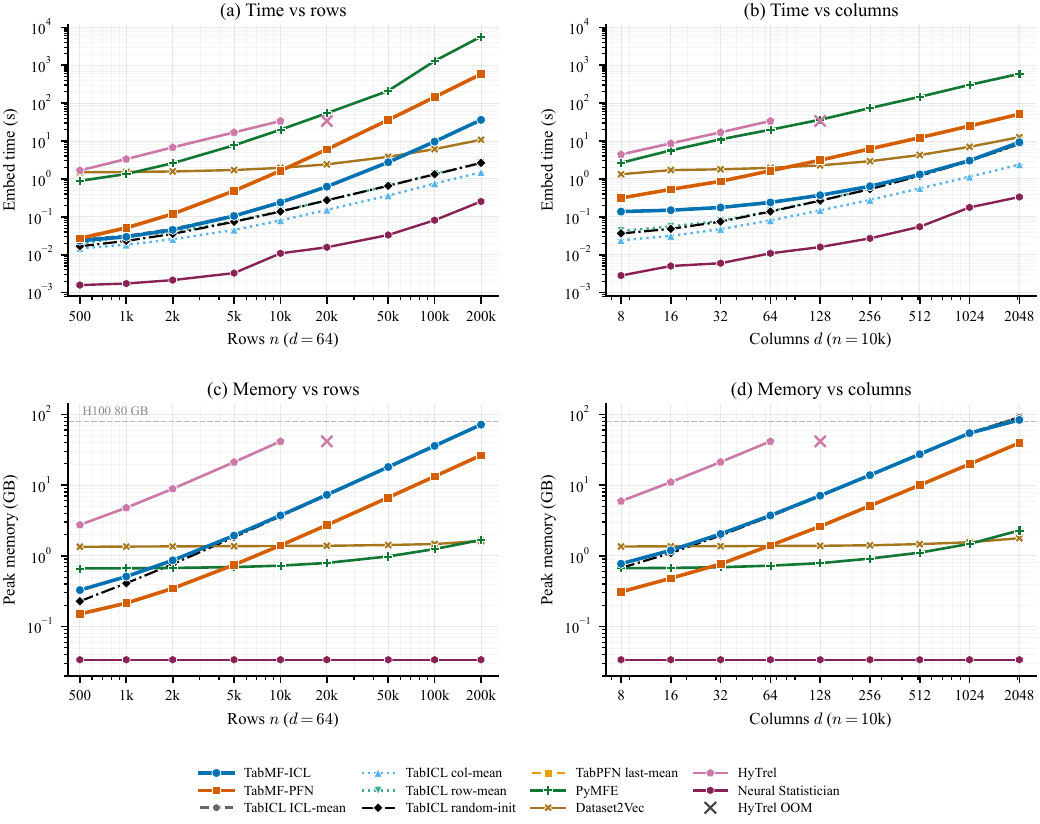}
    \caption{
    Encoder time and memory scaling as the number of rows and features increases.
    For the row-scaling experiment, the number of features is fixed at $d=64$;
    for the feature-scaling experiment, the number of observations is fixed at
    $n=10{,}000$. \methodname{}-ICL closely tracks TabICL ICL-mean, while
    \methodname{}-PFN closely tracks TabPFN mean-pooling, indicating that the
    dataset-level readouts introduce little additional inference cost relative
    to their respective backbones. HyTrel exceeds available memory at larger
    table sizes. Neural Statistician and Dataset2Vec use bounded input sampling,
    which limits their scaling with table size.
    }
    \label{fig:results-efficiency}
\end{figure}

Figure~\ref{fig:results-efficiency} shows embedding time and peak memory.
The main observation is that both \methodname{} variants closely track the
computational cost of their corresponding frozen TFM representations:
\methodname{}-ICL follows TabICL ICL-mean, while \methodname{}-PFN follows
TabPFN mean-pooling. Thus, the dataset-level adaptation and readout introduce
little additional inference cost beyond their respective underlying
backbones. Because Dataset2Vec and the Neural Statistician subsample a bounded input regardless of table size, their constant memory and low embedding time reflect that cap rather than cheaper processing of the full table; TabMF, like its backbone, encodes the entire input.

\begin{table}[t]
\centering
\caption{Supplementary transfer evaluations. TEmBed MAP covers 16
datalakes except PyMFE (\dag: 17); shuffling covers 60 slices.
Standard deviations are across datalakes or slices. Similarity and
shuffling use uncapped inputs. HPO covers 82 targets except HyTrel
(\ddag: 77). A dash denotes an unavailable result.}
\label{tab:results-transfer}
\small
\begin{tabular}{lrrr}
\toprule
Encoder & MAP $\uparrow$ & Shuffling $\uparrow$ & Type F1 $\uparrow$ \\
\midrule
\methodname{}-ICL & $0.605\pm0.158$ & $0.732\pm0.281$ & 0.363 \\
\methodname{}-PFN & $0.448\pm0.117$ & $0.773\pm0.251$ & 0.289 \\
TabICL ICL-mean & $0.468\pm0.137$ & $0.753\pm0.266$ & --- \\
HyTrel & $0.684\pm0.146$ & $0.977\pm0.053$ & 0.681 \\
PyMFE & $0.456\pm0.137^{\dagger}$ & $0.705\pm0.150$ & 0.372 \\
\midrule
Encoder & HPO quality $\uparrow$ & HPO regret@5 $\downarrow$ & Targets \\
\midrule
\methodname{}-ICL & 0.468 & 0.037 & 82 \\
\methodname{}-PFN & 0.676 & 0.032 & 82 \\
TabICL ICL-mean & 0.773 & 0.032 & 82 \\
HyTrel & $0.664^{\ddagger}$ & $0.042^{\ddagger}$ & 77 \\
\bottomrule
\end{tabular}
\end{table}

\clearpage
\section{Illustration of the \methodname{} backbone Aggregations}
\label{sec:illustrationaggregation}

\begin{figure}[h]
\begin{center}
\resizebox{0.98\linewidth}{!}{
\definecolor{figink}{HTML}{343A40}
\definecolor{figgray}{HTML}{868E96}
\definecolor{figfaint}{HTML}{CED4DA}
\definecolor{figlight}{HTML}{F1F3F5}
\definecolor{figacc}{HTML}{82B366}
\definecolor{figacctx}{HTML}{3F6E36}
\definecolor{figaccbg}{HTML}{D5E8D4}
\definecolor{figtar}{HTML}{D79B00}
\definecolor{figtarbg}{HTML}{FFE6CC}

\begin{tikzpicture}[
  every node/.style={
    font=\sffamily\small,
    text=figink,
    inner sep=2pt
  },
  blk/.style={
    draw=figgray,
    fill=figlight,
    rounded corners=3pt,
    line width=0.6pt,
    minimum width=4.0cm,
    minimum height=0.52cm,
    align=center
  },
  ours/.style={
    blk,
    draw=figacc,
    fill=figaccbg,
    text=figacctx
  },
  op/.style={
    draw=figgray,
    fill=white,
    rounded corners=6pt,
    line width=0.5pt,
    minimum width=2.6cm,
    minimum height=0.42cm,
    align=center,
    font=\sffamily\scriptsize
  },
  tarop/.style={
    op,
    draw=figtar,
    text=figtar
  },
  flow/.style={
    -{Latex[length=1.4mm]},
    figink,
    line width=0.55pt
  },
  cont/.style={
    draw=figfaint,
    dashed,
    rounded corners=5pt,
    line width=0.5pt
  },
]


\node at (0,-0.05) {view $(\mathbf{X},\mathbf{y})$};

\draw[flow] (0,0.2) -- (0,0.62);

\node[blk] (l1) at (0,0.9) {
  Column-wise embedding
};

\draw[flow] (l1.north) -- (0,1.5);

\node[blk] (l2) at (0,1.78) {
  Row-wise Transformer
};

\draw[flow] (l2.north) -- (0,2.38);

\node[tarop] (l3) at (0,2.62) {
  $+\ \operatorname{YEnc}(y_t)$
};

\draw[flow] (l3.north) -- (0,3.2);

\node[blk] (l4) at (0,3.48) {
  ICL Transformer $+$ LN
};

\node[
  ours,
  minimum width=1.5cm,
  minimum height=0.42cm,
  font=\sffamily\scriptsize
] (ltok) at (3.2,2.62) {
  $\mathbf{s}_{\mathrm{sum}}$
};

\draw[flow, figacc]
  (ltok.north) |- (1.1,3.05) -- (1.1,3.22);

\node[
  font=\sffamily\scriptsize,
  text=figacctx,
  align=center
] at (3.2,2.1) {
  summary token\\
  (learned)
};

\draw[cont]
  (-2.35,0.58) rectangle (2.35,3.86);

\node[
  anchor=west,
  font=\sffamily\scriptsize,
  text=figgray
] at (-2.3,4.02) {
  TabICLv2 backbone (pretrained)
};

\draw[flow] (l4.north) -- (0,4.38);

\node[op] (l5) at (0,4.62) {
  take position $T{+}1$
};

\draw[flow] (l5.north) -- (0,5.18);

\node[ours] (l6) at (0,5.46) {
  Projection $\projector$ $+$ $\ell_2$
};

\draw[flow] (l6.north) -- (0,6.02);

\fill[figaccbg]
  (-0.09,6.08) rectangle (0.09,6.72);

\foreach \i in {1,2,3}{
  \draw[
    figacc,
    line width=0.4pt
  ]
  (-0.09,6.08+0.16*\i)
  --
  (0.09,6.08+0.16*\i);
}

\draw[
  figacc,
  line width=0.5pt
]
(-0.09,6.08) rectangle (0.09,6.72);

\node[text=figacctx] at (0.42,6.4) {
  $\embedding$
};

\node[
  font=\sffamily\bfseries\small,
  text=figink
] at (0,7.05) {
  \methodname{}-ICL
};


\node at (8.6,-0.05) {
  view $(\mathbf{X},\mathbf{y})$
};

\draw[flow] (8.6,0.2) -- (8.6,0.62);

\node[blk] (r1) at (8.6,0.9) {
  Feature-wise attention
};

\draw[flow] (r1.north) -- (8.6,1.5);

\node[blk] (r2) at (8.6,1.78) {
  Sample-wise attention
};

\draw[
  figgray,
  line width=0.5pt
]
(6.35,0.62)
--
(6.2,0.62)
--
(6.2,2.06)
--
(6.35,2.06);

\node[
  font=\sffamily\scriptsize,
  text=figgray
] at (5.95,1.34) {
  $L\times$
};

\draw[cont]
  (6.55,0.58) rectangle (10.95,2.14);

\node[
  anchor=west,
  font=\sffamily\scriptsize,
  text=figgray
] at (6.6,2.32) {
  TabPFN-2.5 backbone (pretrained)
};

\draw[flow] (r2.north) -- (8.6,2.66);

\node[op] (r3) at (8.6,2.9) {
  mean over observations
};

\node[
  anchor=west,
  font=\sffamily\scriptsize,
  text=figgray,
  align=left
] at (10.0,2.9) {
  labeled rows only;\\
  target column and\\
  thinking rows dropped
};

\draw[flow] (r3.north) -- (8.6,3.46);

\node[ours] (r4) at (8.6,3.74) {
  Column Transformer $\operatorname{TF}_{\mathrm{col}}$
};

\draw[flow] (r4.north) -- (8.6,4.34);

\node[op] (r5) at (8.6,4.58) {
  masked mean over groups
};

\draw[flow] (r5.north) -- (8.6,5.18);

\node[ours] (r6) at (8.6,5.46) {
  Projection $\projector$ $+$ $\ell_2$
};

\draw[flow] (r6.north) -- (8.6,6.02);

\fill[figaccbg]
  (8.51,6.08) rectangle (8.69,6.72);

\foreach \i in {1,2,3}{
  \draw[
    figacc,
    line width=0.4pt
  ]
  (8.51,6.08+0.16*\i)
  --
  (8.69,6.08+0.16*\i);
}

\draw[
  figacc,
  line width=0.5pt
]
(8.51,6.08) rectangle (8.69,6.72);

\node[text=figacctx] at (9.02,6.4) {
  $\embedding$
};

\node[
  font=\sffamily\bfseries\small,
  text=figink
] at (8.6,7.05) {
  \methodname{}-PFN
};


\fill[
  figlight,
  draw=figgray,
  line width=0.5pt
]
(1.0,-0.85) rectangle (1.3,-0.65);

\node[
  anchor=west,
  font=\sffamily\scriptsize,
  text=figgray
] at (1.35,-0.75) {
  pretrained, fine-tuned
};

\fill[
  figaccbg,
  draw=figacc,
  line width=0.5pt
]
(4.6,-0.85) rectangle (4.9,-0.65);

\node[
  anchor=west,
  font=\sffamily\scriptsize,
  text=figgray
] at (4.95,-0.75) {
  added, trained from scratch
};

\draw[
  figtar,
  line width=0.6pt,
  rounded corners=2pt
]
(8.8,-0.85) rectangle (9.1,-0.65);

\node[
  anchor=west,
  font=\sffamily\scriptsize,
  text=figgray
] at (9.15,-0.75) {
  target pathway
};

\end{tikzpicture}}
\end{center}
\caption{
The two \methodname{} encoders.
\methodname{}-ICL (left) appends a learned summary token to the
target-conditioned row states and reads its state after the pretrained
ICL Transformer and final normalization.
\methodname{}-PFN (right) pools the backbone's feature-group states over
labeled observations, models their interactions with a learned column
Transformer, and averages the active groups.
Each readout feeds a projection network; gray blocks are pretrained and
fine-tuned, while green blocks are added and trained from scratch.
}
\label{fig:encoders}
\end{figure}
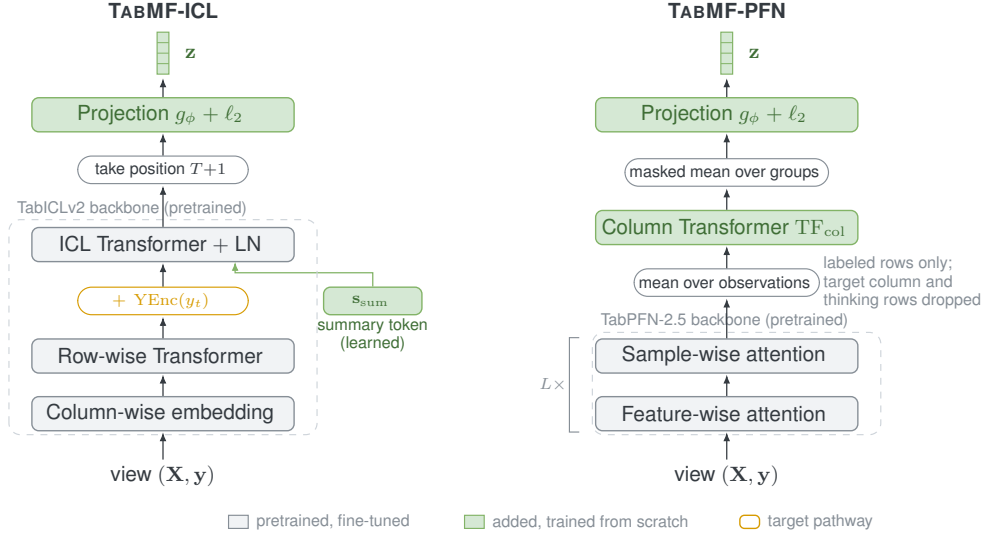

\end{document}